%% file: main.tex
\documentclass[10pt,twocolumn,letterpaper]{article}

\PassOptionsToPackage{table}{xcolor}

\usepackage{cvpr}              % To produce the CAMERA-READY version
\definecolor{cvprblue}{rgb}{0.21,0.49,0.74}
\usepackage[pagebackref,breaklinks,colorlinks,allcolors=cvprblue]{hyperref}

\def\paperID{18131} % *** Enter the Paper ID here
\def\confName{CVPR}
\def\confYear{2025}

\usepackage{glossaries}
\usepackage{bbm}

\definecolor{lightgray}{gray}{0.9}
\definecolor{linecolor}{rgb}{0.82, 0.94, 0.75}

\hypersetup{
    colorlinks=true,
    linkcolor=red,
    filecolor=magenta,      
    urlcolor=magenta,
}

\usepackage{multirow} 

\usepackage{glossaries}

\newacronym{tta}{TTA}{Test-Time Adaptation}
\newacronym{ttt}{TTT}{Test-Time Training}
\newacronym{knn}{KNN}{K-Nearest Neighbors}
\newacronym{mae}{MAE}{Masked Autoencoder}
\newacronym{fps}{FPS}{Farthest Point Sampling}
\newacronym{bn}{BN}{Batch Normalization}

\usepackage{graphicx}

\newcommand{\mypar}[1]{\vspace{2.5pt}
\noindent\textbf{#1~}}

\newcommand{\improv}[1]{\ \tiny \textcolor{ForestGreen}{(+#1\%)}}
\newcommand{\decr}[1]{\ \tiny \textcolor{BrickRed}{(-#1\%)}}

\colorlet{MyCol}{Tan!10}

\newcommand{\colrow}{\rowcolor{MyCol}}

\newcommand{\phz}{\phantom{0}}

\newcommand{\rotb}[1]{\rotatebox{70}{#1}}
\title{SMART-PC \\ Skeletal Model Adaptation for Robust Test-Time Training in Point Clouds}

\author{
Ali Bahri \thanks{Correspondence to \href{mailto:ali.bahri.1@ens.etsmtl.ca}{ali.bahri.1@ens.etsmtl.ca}} \and 
Moslem Yazdanpanah \and
Sahar Dastani \and 
Mehrdad Noori \and  
Gustavo Adolfo Vargas Hakim \and  
David Osowiechi \and  
Farzad Beizaee \and
Ismail Ben Ayed \and 
Christian Desrosiers \\
\\
LIVIA, ÉTS Montréal, Canada\\
International Laboratory on Learning Systems (ILLS)\\
}

\begin{document}
\maketitle
\input{sec/0_abstract}    
\input{sec/1_intro}
\input{sec/2_related_work}
\input{sec/3_method}

{
    \small
    \newpage
    \bibliographystyle{ieeenat_fullname}
    \bibliography{main}
}

\input{sec/supp}

\end{document}

%% file: sec/0_abstract.tex
\begin{abstract}
\acrfull{ttt} has emerged as a promising solution to address distribution shifts in 3D point cloud classification. However, existing methods often rely on computationally expensive backpropagation during adaptation, limiting their applicability in real-world, time-sensitive scenarios. In this paper, we introduce SMART-PC, a skeleton-based framework that enhances resilience to corruptions by leveraging the geometric structure of 3D point clouds. During pre-training, our method predicts skeletal representations, enabling the model to extract robust and meaningful geometric features that are less sensitive to corruptions, thereby improving adaptability to test-time distribution shifts.
Unlike prior approaches, SMART-PC achieves real-time adaptation by eliminating backpropagation and updating only BatchNorm statistics, resulting in a lightweight and efficient framework capable of achieving high frame-per-second rates while maintaining superior classification performance. Extensive experiments on benchmark datasets, including ModelNet40-C, ShapeNet-C, and ScanObjectNN-C, demonstrate that SMART-PC achieves state-of-the-art results, outperforming existing methods such as MATE in terms of both accuracy and computational efficiency. The implementation is available at: \url{https://github.com/AliBahri94/SMART-PC}.

\end{abstract}

%% file: sec/1_intro.tex
\section{Introduction}
\label{introduction}

% Recent advancements in deep learning have led to remarkable breakthroughs in processing and classifying 3D point clouds. Despite these achievements, the performance of these models often relies on the assumption that the test data distribution matches the training data distribution. 
% However, when the test distribution (target) deviates from the training distribution (source), the resulting distribution shift poses significant challenges for machine learning systems deployed in real-world settings. For example, point cloud data captured by sensors such as LiDAR can be heavily impacted by environmental conditions, sensor inaccuracies, or other external factors, leading to notable distortions.
% Studies \cite{ren2022benchmarking, sun2022benchmarking} have demonstrated that even minor disturbances, such as small-scale noise or geometric distortions, can substantially degrade the performance of state-of-the-art 3D recognition models. This vulnerability poses challenges in critical domains such as autonomous navigation, urban mapping, industrial automation, and robotics. Given the diversity and unpredictability of distributional shifts in 3D data, it is impractical to pre-train models for all possible scenarios. Consequently, there is a growing need for methods that enable robust, unsupervised adaptation to distributional changes directly at test time.

Recent advancements in deep learning have significantly improved the classification of 3D point clouds \cite{pang2022masked, zhang2022point, si-mamba, liang2024pointmamba, geomask3d}. However, these models often assume that the test data distribution matches the training distribution, an assumption that rarely holds in real-world scenarios. Distribution shifts, caused by factors such as environmental conditions or sensor inaccuracies in LiDAR data, can introduce distortions that degrade model performance \cite{ren2022benchmarking, sun2022benchmarking}. This vulnerability is particularly challenging in critical domains like autonomous navigation and robotics. As pre-training for all potential scenarios is impractical, robust methods for unsupervised test-time adaptation are essential to address these shifts effectively.

\begin{figure}
	\centering
\includegraphics[width=0.9 \linewidth]{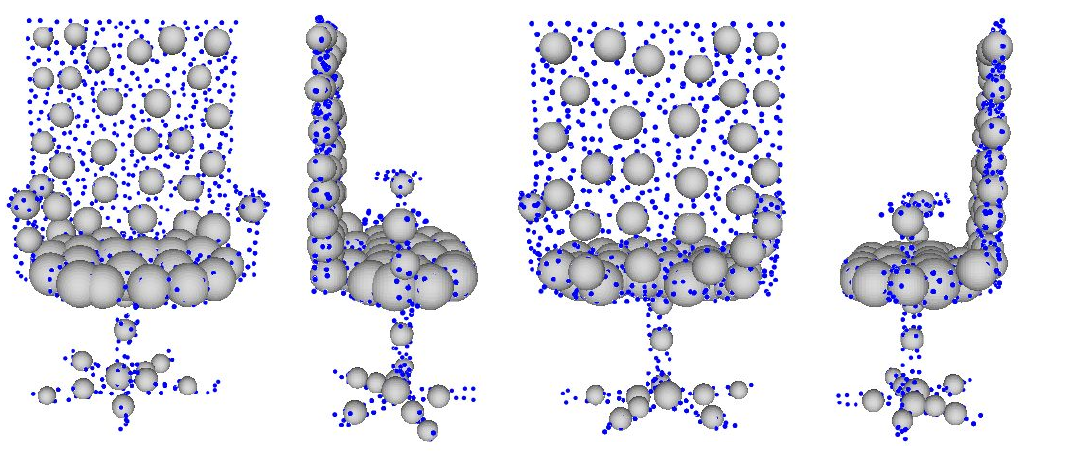}
\vspace{-15pt}

	\caption{The blue points represent the sampled points on the surface of spheres created using the skeletal points as centers and their corresponding radii. Each sphere illustrates the local geometric structure defined by the skeletal representation.}
	\label{fig:chair_2}
    \vspace{-15pt}
\end{figure}

In \gls{ttt} for classification, an additional technique is employed during pre-training with the source dataset, which can be leveraged during the adaptation phase. During adaptation, the network utilizes unlabeled target data to dynamically adjust to shifts in data distributions at test time. Recent \gls{ttt} approaches in the 2D image domain have explored various strategies, such as entropy-based regularization, updating BatchNorm statistics, and self-supervised tasks \cite{liang2020we, wang2020tent, mirza2022norm, sun2020test}. However, these methods struggle when directly applied to 3D point clouds, highlighting the need for \gls{ttt} techniques specifically designed for 3D data. 
In the 3D domain, a notable work is the MATE framework \cite{mirza2023mate}, which addresses the problem of \gls{ttt} for 3D point cloud classification. MATE introduces a 3D-specific approach that leverages the self-supervised paradigm, where a network is adapted to out-of-distribution (OOD) target data by solving a self-supervised task. 
MATE utilizes a masked autoencoder (MAE), where the network reconstructs a point cloud from its partially removed data. This reconstruction process enables adaptation, enhancing robustness to distribution shifts during test-time.

% While MATE represents a significant advancement in \gls{ttt} for 3D point clouds, it has critical limitations. The reliance on a masked autoencoder (MAE) for pre-training, while effective for reconstruction tasks, does not enable the extraction of highly meaningful or powerful geometric features that are essential for robust and generalizable representations. This limitation stems from the reconstruction objective’s emphasis on restoring missing parts of the input, which may not fully leverage the inherent structural and geometric properties of 3D point clouds for robust feature learning. As a result, MATE heavily depends on updating all network parameters during adaptation to compensate for the lack of robust features learned during pre-training. This approach reduces the model's speed and makes it less practical for real-time applications due to its lower Frames/Second.

While MATE represents a significant advancement in \gls{ttt} for 3D point clouds, it has critical limitations. MATE focuses on reconstructing points on the surface of objects, making it highly sensitive to surface-level corruptions, such as noise, which frequently occur in real-world scenarios. 
% The reliance on a masked autoencoder (MAE) for pretraining, while effective for reconstruction tasks, does not enable the extraction of meaningful geometric features that capture the global structure of 3D objects. 
This focus on surface points makes the features learned by MATE highly sensitive to corruptions, resulting in reduced resilience to distribution shifts. Additionally, MATE heavily depends on updating all network parameters during adaptation. This approach reduces the model's speed, making it less practical for real-time applications due to its lower Frames/Second.

% This approach not only increases computational overhead but also reduces the efficiency and practicality of the method, particularly for real-time applications.

% In contrast, our method overcomes these limitations by introducing a skeleton-based framework for \gls{ttt}. Skeletons inherently capture the essential geometric structure of 3D point clouds, allowing the model to learn powerful and meaningful features that are less sensitive to corruptions during pre-training.
In contrast, our method overcomes these limitations by introducing a skeleton-based framework for \gls{ttt}. By capturing the essential geometric structure of 3D point clouds, skeletons enable the model to learn powerful and meaningful features that are inherently less sensitive to corruptions. This is achieved through the abstraction property of skeletal representations \cite{lin2021point2skeleton}, which simplifies the shape into a compact form, filtering out high-frequency noise and local distortions. As a result, the model focuses on the underlying geometric structure rather than surface-level variations, enhancing robustness to distribution shifts.
By leveraging these robust features, our approach eliminates the need for backpropagation during adaptation, significantly improving speed. Instead, following prior studies \cite{nado2020evaluating, li2018adaptive}, we update only the BatchNorm statistics, which effectively adapts the model to distribution shifts without modifying its learned weights. 
This efficiency demonstrates that skeleton-based pre-training inherently equips the model with features that are more resilient to corruption compared to MATE, providing a lightweight and effective solution for \gls{ttt}. As shown in \Cref{fig:chair_2}, the skeletal representation captures the geometric structure of point clouds by abstracting them into compact and meaningful features.
Our method not only overcomes the limitations of MATE but also introduces a paradigm shift in \gls{ttt} by emphasizing the importance of meaningful feature extraction during pre-training, ensuring robustness and adaptability with minimal computational effort. Our main contributions are:
\begin{itemize}[noitemsep,topsep=0pt]
    % \item We propose SMART-PC, a skeleton-based pre-training framework for \gls{ttt} that enables the network to extract meaningful and robust geometric features, enhancing the model's resistance to distribution shifts.
    \item We propose SMART-PC, a skeleton-based framework for \gls{ttt} that enhances robustness by extracting geometric features that are less sensitive to distribution shifts.
    % \item Our method achieves test-time adaptation without backpropagation, relying solely on updating BatchNorm statistics, significantly reducing computational overhead. 
    \item Our method achieves test-time adaptation without backpropagation, relying solely on updating BatchNorm statistics, making it highly efficient for real-time applications.

    \item We achieve state-of-the-art results on benchmark datasets, demonstrating the effectiveness of our approach compared to existing methods, including MATE.
\end{itemize}

%% file: sec/2_related_work.tex
\section{Related Works}

\mypar{Test-Time Training.} \gls{ttt} enhances model adaptability by leveraging unseen target data during inference, unlike domain generalization or adaptation methods, which operate solely during training. \gls{ttt} methods can be broadly categorized into regularization-based and self-supervised approaches. Regularization-based methods include TENT \cite{wang2020tent}, which minimizes prediction entropy by adapting BatchNorm parameters, and SHOT \cite{liang2020we}, which combines entropy minimization with diversity regularization to develop robust feature extraction. MEMO \cite{zhang2022memo} applies data augmentations to test inputs and minimizes the entropy of averaged outputs, improving robustness. DUA \cite{mirza2022norm} updates BatchNorm statistics to address distribution shifts with minimal overhead, while T3A \cite{iwasawa2021test} refines the linear classifier using pseudo-prototypes, achieving backpropagation-free adaptation. These methods highlight the diversity of \gls{ttt} strategies for addressing distribution shifts.

Among self-supervised approaches, TTT++ \cite{liu2021ttt++} introduces an additional self-supervised branch that leverages contrastive learning within the source model to facilitate adaptation to the target domain. Another method, TTT-MAE \cite{gandelsman2022test}, employs a \gls{mae} to address the one-sample learning problem effectively, demonstrating improved performance across various visual benchmarks.
A recent method, MATE \cite{mirza2023mate}, is the first \gls{ttt} framework specifically tailored for 3D point cloud data, enhancing the resilience of deep networks to distribution shifts during testing. It leverages a \gls{mae} objective, masking a significant portion of each target point cloud and tasking the network with reconstructing the complete structure before classification. In addition to these approaches, several works on test-time adaptation have explored updating model parameters during inference to handle distribution shifts effectively \cite{wang2024backpropagation, bahri2024test}. 
%Finally, BFTT3D \cite{wang2024backpropagation} introduces a backpropagation-free \gls{tta} method specifically designed for 3D data, addressing domain shifts with a two-stream architecture that maintains both source and target domain knowledge.

\mypar{Skeletal Representation.}
Skeletal representations are a compact and structural abstraction of 3D point clouds, capturing the underlying geometric and topological properties of objects. These representations simplify complex point cloud data by reducing it to a set of key skeletal points that capture the core of the geometry (\cref{fig:chair_2}), which are particularly useful for tasks such as shape analysis, object recognition, and reconstruction. 
Several methods have been proposed to generate skeletal representations from 3D data. Early approaches often relied on medial axis transformations (MAT) \cite{choi1997mathematical} or Voronoi-based \cite{ogniewicz1992voronoi} methods to extract the central skeleton of an object. While effective in capturing global geometry, these methods are typically computationally intensive and sensitive to noise \cite{li2015q, sun2015medial, yan2018voxel}. 
 % \CD{Reference(s) supporting this?}. 
More recent learning-based approaches, such as Point2Skeleton \cite{lin2021point2skeleton} and Learnable Skeleton-Aware 3D Point Cloud Sampling \cite{wen2023learnable}, have shifted towards using neural networks to predict skeletal points directly from input point clouds. These methods leverage local geometric features and end-to-end optimization to produce accurate and robust skeletal representations. By capturing the core geometric structure of 3D point clouds, skeletal representations provide a compact and meaningful abstraction that is less sensitive to noise and distortions compared to raw point cloud data. This robustness makes them particularly suitable for dynamic adaptation scenarios, where models need to generalize effectively to unseen conditions. In this paper, we leverage skeletal representations with technical enhancements to enable the network to learn robust geometric features, thereby eliminating the need for extensive parameter updates during \gls{tta}. Instead, our method relies solely on lightweight adjustments, such as updating BatchNorm statistics. This approach ensures both efficient and effective \gls{ttt} while fully utilizing the inherent strengths of skeletal representations.

\begin{figure*}[!t]
\centering
\includegraphics[width=0.9\textwidth]{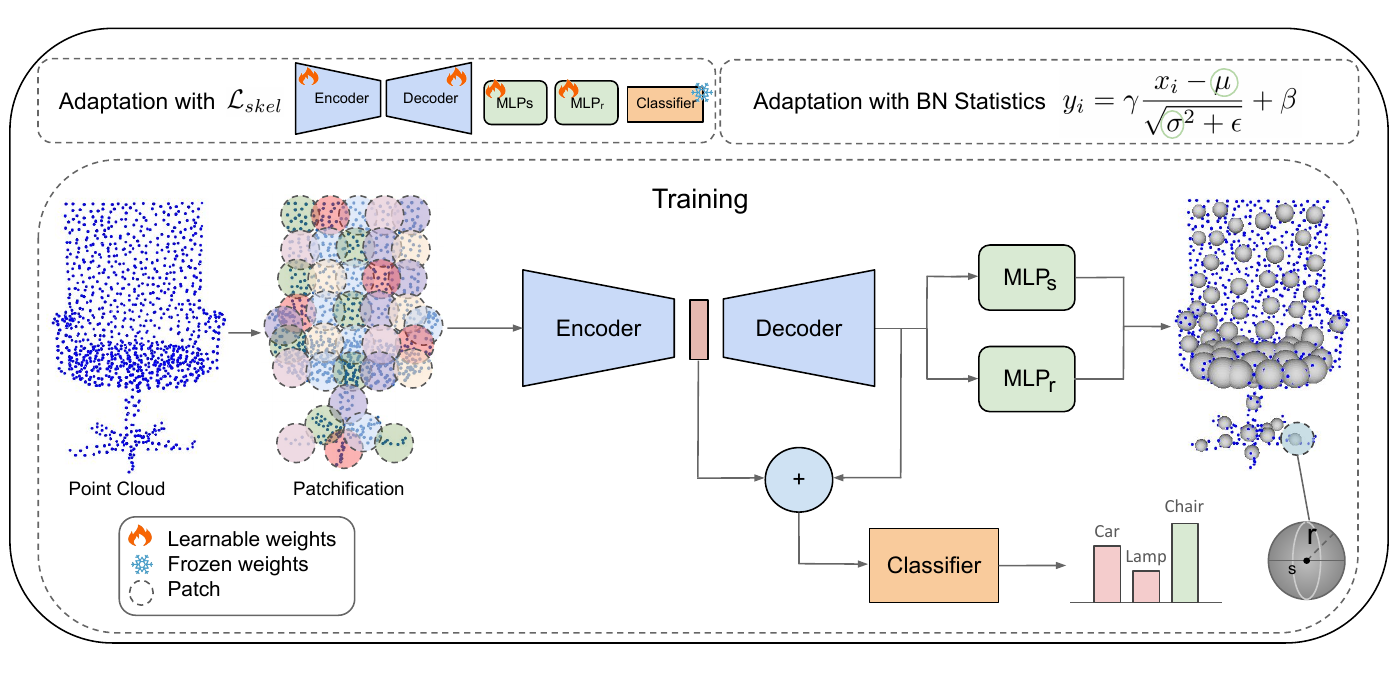}

\vspace{-12pt}
\caption{An overview of the SMART-PC framework. The framework integrates skeletal prediction and classification tasks, leveraging skeletal representations to extract robust geometric features. During online adaptation, two strategies are employed: adaptation with the skeletal loss $\mathcal{L}_{skel}$ and backpropagation, and a lightweight, backpropagation-free approach that updates only BatchNorm statistics. 
%This design ensures efficient and effective adaptation to distribution shifts, making SMART-PC suitable for real-time applications with higher frame rates.
}
\label{main}
\end{figure*} 

%% file: sec/3_method.tex
\section{Method}

\subsection{Preliminaries}

A point cloud is represented as $P \in \mathbb{R}^{N \times 3}$, where $N$ is the total number of points, and each point is defined by its 3D spatial coordinates $(x, y, z)$. This representation captures the geometric structure of 3D objects, but due to the irregular and unordered nature of point clouds, processing all $N$ points can be computationally expensive and redundant.
To reduce redundancy and focus on the most representative points, we apply \gls{fps}. This technique selects a subset of $M$ points from the original point cloud as \textit{centers}, denoted as:
\begin{equation}
C = \mathit{FPS}(P) \in \mathbb{R}^{M \times 3},
\end{equation}
where $M \ll N$. The FPS algorithm ensures that the selected centers are evenly distributed across the point cloud, preserving the overall geometric structure.
For each center point $c_i \in C$, we construct its local neighborhood by selecting its $K$ nearest neighbors from the original point cloud $P$ using the \gls{knn} algorithm. This results in a neighborhood tensor:
\begin{equation}
P_{local} = \mathit{kNN}(C, P) \in \mathbb{R}^{M \times K \times 3},
\end{equation}
where $K$ is the number of neighbors for each center point.
Each neighborhood $P_{local}[i] \in \mathbb{R}^{K \times 3}$, corresponding to a center $c_i$, captures the local geometric features around $c_i$. These neighborhoods provide localized context, allowing the network to efficiently process the point cloud by focusing on both global structure (via the centers) and local geometric details (via the neighbors). This hierarchical tokenization forms the basis for subsequent feature extraction and skeletal prediction.

With the point cloud tokenized into centers and their local neighborhoods, the next step is to define the skeletal representation. The skeleton of a 3D shape provides a compact representation of its intrinsic geometric structure. It abstracts the shape by representing it as a collection of skeletal points and their corresponding radii, which together form a \textit{skeletal mesh}. This concept was introduced by \cite{lin2021point2skeleton} for learning-based approaches.
A skeletal mesh is defined as a discrete set of skeletal spheres. Each sphere is represented as:
\begin{equation}
s = \big(c_s, r(c_s)\big) \in \mathbb{R}^4,
\end{equation}
where $c_s \in \mathbb{R}^3$ is the center of the skeletal sphere, referred to as the \textit{skeletal point}, and $r(c_s) \in \mathbb{R}$ %$r(c_s): \mathbb{R}^3 \to \mathbb{R}$ 
is its associated radius, representing the distance from the center $c_s$ to the surface of the original 3D shape. 
The skeletal mesh is constructed by connecting the skeletal spheres using two main elements. First, \emph{edges}, denoted as $e_{ij} = (c_i, c_j)$, connect two skeletal points $c_i$ and $c_j$, forming the basic linear structure of the skeleton. Second, \emph{faces}, denoted as $f_{ijk} = (c_{s_{i}}, c_{s_{j}}, c_{s_{k}})$, are triangular connections between three skeletal points that capture 2D non-manifold surfaces within the mesh.
% \CD{Having a figure showing this might be useful.}.

The skeletal representation possesses two key properties. The first is \textbf{recoverability}, where the skeletal mesh acts as a complete descriptor of the shape. This means the original 3D shape can be reconstructed by interpolating the skeletal spheres. For a skeletal sphere $s = (c_s, r(c_s))$, any point $p$ on its surface can be expressed as:
\begin{equation}
p \, = \, c_s \, + \, r(c_s) \cdot v,
\end{equation}
where $v$ is a unit vector specifying the direction.
The second property is \textbf{abstraction}, by which the skeleton simplifies the representation of the shape by capturing its fundamental structure, significantly reducing the complexity of the original point cloud while retaining its essential geometric information.
This discrete skeletal representation approximates the Medial Axis Transform (MAT), which is a continuous formulation of the skeleton. Unlike the MAT, the skeletal mesh is more robust to boundary noise and perturbations, making it suitable for learning-based approaches. By predicting skeletal points and their radii, the skeletal representation provides a robust and efficient way to analyze and process 3D shapes, even under noisy or corrupted conditions.

\subsection{Overview}

In this paper, we propose a novel skeleton-based framework for \acrfull{ttt} designed to improve the robustness and adaptability of 3D point cloud classification under distribution shifts. Our method consists of three components: \textit{Framework}, \textit{Training}, and \textit{Test-Time Adaptation} (\acrshort{tta}).

% In the \textit{Framework} phase, we define the structure of the model, which includes components for both classification and skeletal point prediction. This phase involves investigating the architectural details and ensuring the network is capable of handling both tasks effectively \CD{This is a bit unusual for a phase, where a temporal aspect is implied. Do you perhaps mean a component?}.

In the \textit{Framework} component, we define the structure of the model, which includes components for both classification and skeletal point prediction. This involves investigating the architectural details and ensuring the network is capable of handling both tasks effectively.
During the \textit{Training} phase, the model is trained on source data to perform two tasks: skeletal prediction and classification. The skeletal-based pre-training enables the model to learn robust and meaningful geometric features that enhance resistance to corruptions and generalize effectively across varying distributions.
In the \acrshort{tta} phase, our method eliminates the need for backpropagation when the adaptation strategy is online, meaning the model is not reset after each epoch.
Instead, it relies solely on lightweight updates to the BatchNorm statistics, enabling the model to adapt dynamically to target data while achieving a higher frame rate for real-time applications.
% By leveraging the inherent robustness of skeletal representations, our approach achieves state-of-the-art results in both ``source only'' and adaptation scenarios, demonstrating its effectiveness and scalability in real-world 3D data applications.
The overall network architecture is illustrated in \cref{main}.

\subsection{Framework}

Our framework is designed to perform two primary tasks: skeletal point and radius prediction, and point cloud classification. These tasks share a unified encoder while leveraging separate network components for their specific objectives. The framework is divided into two key branches: the \textit{skeletal branch} for predicting skeletal representations and the \textit{classification branch} for predicting class labels.

\mypar{Skeletal Branch.} 
The skeletal branch predicts skeletal points $S \in \mathbb{R}^{M \times 3}$ and their corresponding radii $R \in \mathbb{R}^{M \times 1}$ from the input point cloud. The encoder, denoted as $E$, processes the tokenized point cloud $P_{local} \in \mathbb{R}^{M \times K \times 3}$ to extract global and local geometric features:
\begin{equation}
F_{enc} = E(P_{local}) \in \mathbb{R}^{M \times d},
\end{equation}
where $d$ is the dimensionality of the feature space. The encoder captures both local geometric details and global structural information.

The decoder, denoted as $D$, refines the encoded features to produce a contextually enriched representation:
\begin{equation}
F_{dec} = D(F_{enc}) \in \mathbb{R}^{M \times d}.
\end{equation}

To predict skeletal points and radii, two separate Multi-Layer Perceptrons ($MLP_s$ and $MLP_r$) are applied to the refined features $F_{dec}$. The skeletal points are predicted as:
\begin{equation}
c_s = \mathit{MLP}_{s}(F_{dec}) \in \mathbb{R}^{M \times 3},
\end{equation}
and the radii are predicted as:
\begin{equation}
r = \mathit{MLP}_{r}(F_{dec}) \in \mathbb{R}^{M \times 1}.
\end{equation}

The skeletal branch leverages these predictions to model the underlying geometric structure of the input point cloud. Unlike existing skeletal prediction methods \cite{lin2021point2skeleton, wen2023learnable}, which rely on convex combinations of input points to generate skeletal points, our approach directly predicts the skeletal points and radii from the refined features $ F_{dec}$. Convex combination methods inherently inject raw point cloud data into the prediction process, making the task easier for the network. However, this dependency on the raw point cloud can hinder the encoder's ability to learn robust and meaningful geometric features, especially in the presence of noise or distribution shifts.

\mypar{Classification Branch.}
The classification branch uses the shared encoder to extract features for class prediction. To enhance the classification performance, the features from the encoder $F_\text{enc}$ and decoder $F_\text{dec}$ are combined by a sum:
\begin{equation}
F_{\mathit{combined}} = F_{enc} + F_{dec}.
\end{equation}
This combination allows the classification head to leverage high-level features from the decoder that are closely related to the skeletal points, alongside the global context from the encoder. By incorporating these skeletal-related features, the model enhances its ability to classify point clouds, particularly by utilizing structural information that aids in distinguishing complex or similar classes.
The combined features are passed through the classification head, which consists of multiple MLP layers, normalization, and dropout layers, to predict class probabilities:
\begin{equation}
p = \mathit{Softmax}(\mathit{MLP}_{\!cls}(F_{\mathit{combined}})) \in \mathbb{R}^K,
\end{equation}
where $K$ is the number of classes.

% The skeletal branch and classification branch work together to enable the network to handle both structural representation and class prediction effectively. The shared encoder not only reduces computational overhead but also ensures that the learned features are jointly optimized for both tasks, making the framework efficient and scalable.

\subsection{Training}
\label{training}
In our framework, the skeletal branch is trained in a self-supervised manner since no labels are available for skeletal points, while the classification branch is trained in a supervised manner using labeled source data. To optimize these branches, we adopt loss functions inspired by prior works in skeletal representation~\cite{lin2021point2skeleton, wen2023learnable}. Below, we detail the loss functions used for training the skeletal and classification branches.

\mypar{Skeletal Losses.} 
The skeletal branch optimizes three complementary loss functions to ensure accurate prediction of skeletal points and their corresponding radii. Unlike existing skeletal methods that operate at the point level, our approach predicts skeletal points and radii at the patch level. For each patch, the model predicts the center of a skeletal sphere and its radius, which together abstract the local structure of the point cloud.

\textbf{1) Point-to-Sphere Loss.} This loss ensures that input points lie on the surface of their corresponding skeletal spheres and that skeletal spheres align closely with their associated input points $P$:
\begin{align}
\mathcal{L}_{{p2s}} = &\sum_{p \in P} \left( \min_{s \in S} \|p - c_s\|_2 - r(c_s) \right) \notag \\[-4pt]
&\ \ + \sum_{s \in S} \left( \min_{p \in P} \|c_s - p\|_2 - r(c_s) \right),
\end{align}

% \begin{equation}
% \mathcal{L}_{p2s} = \sum_{p \in P} \left( \min_{s \in S} \|p - c_s\|_2 - r(c_s) \right) + \sum_{s \in S} \left( \min_{p \in P} \|c_s - p\|_2 - r(c_s) \right),
% \end{equation}
where $c_s$ and $r(c_s)$ again denote the center and radius of skeletal sphere $s$.

\textbf{2) Sampling Loss.} To further ensure alignment between the skeletal spheres and the input point cloud, the Chamfer Distance between sampled points on the surface of skeletal spheres and the input points is calculated:
\begin{equation}
\mathcal{L}_{\mathit{sampling}} = \sum_{p \in P} \min_{t \in T} \|p - t\|_2 + \sum_{t \in T} \min_{p \in P} \|t - p\|_2,
\end{equation}
where $T$ represents points sampled uniformly on the surfaces of the skeletal spheres. This loss aggregates geometric information from the neighborhood of input points, effectively reducing high-frequency noise. As each skeletal point represents the local center of a region in $P$, the influence of individual noisy points is averaged, resulting in $s$ being less sensitive to corruption.

\textbf{3) Radius Regularization Loss.} To avoid the instability caused by overly small radii due to noise, a regularization term that encourages larger radii is also used:
\begin{equation}
\mathcal{L}_{\mathit{radius}} = -\sum_{s \in S} r(c_s).
\end{equation}
Furthermore, the radii $r$, predicted alongside $c_s$, capture the extent of each skeletal point's influence. This loss encourages larger radii, reducing sensitivity to localized noise by ensuring that each skeletal sphere represents a broader region of the point cloud. This abstraction minimizes the effect of outliers, further enhancing robustness.

% \CD{This loss makes me uncomfortable. What stops the network from simply predicting infinitely large radii to minimize its objective? Should we cap the loss (e.g., $\sum_s \max(r(c_s), \tau_{\max})$) or apply some monotone transform (e.g., $\sum_s \log r(c_s)$)?}

The total skeletal loss is a weighted combination of these three terms:
\begin{equation}
\mathcal{L}_{skel} \, = \, \mathcal{L}_{p2s} + \lambda_1 \mathcal{L}_{\mathit{sampling}} + \lambda_2 \mathcal{L}_{\mathit{radius}},
\end{equation}
where $\lambda_1 \geq 0$ and $\lambda_2 \geq 0$ are hyperparameters balancing the contributions of each loss.

\mypar{Classification Loss.} 
For the classification branch, we train the network using labeled source data. The features from the encoder and decoder are combined as described earlier and passed to the classification head. The classification loss is defined as the cross-entropy between the predicted probabilities and the ground truth labels:
\begin{equation}
\mathcal{L}_{cls} = -\frac{1}{B} \sum_{i=1}^B \sum_{k=1}^{K} y_{ik} \log(\hat{y}_{ik}),
\end{equation}
where $B$ is the batch size, $K$ is the number of classes, $y_{ik}$ is the ground truth label for the $i$-th sample in class $k$, and $\hat{y}_{ik}$ is the predicted probability for the same class.

The overall loss for training the network is a combination of the skeletal and classification losses:
\begin{equation}
\mathcal{L}_{total} = \mathcal{L}_{skel} + \mathcal{L}_{cls}.
\end{equation}
This joint optimization allows the model to learn robust skeletal representations in a self-supervised manner while simultaneously leveraging labeled source data for classification.

\subsection{\acrfull{tta}}

In this work, we explore two modes of test-time adaptation, inspired by prior works such as MATE~\cite{mirza2023mate}: \textit{online adaptation} and \textit{standard adaptation}. Both modes aim to adapt the model to corrupted target data during test time but differ in how the model state is managed across epochs and corruptions.

\mypar{Online Adaptation.}
In the online mode, the model is not reset after each epoch, allowing the accumulated information from previous batches to influence the adaptation process. However, the model is reset at the beginning of adaptation for each new corruption type.
Initially, we perform test-time adaptation in a backpropagation-free manner, updating only the statistical parameters of the Batch Normalization (BatchNorm) layers (i.e., the running mean and running variance). This approach leverages the observation that the features extracted by the model during pre-training on source data are highly robust and less sensitive to corruption. By updating only the BatchNorm statistics, the model dynamically adjusts to the target data, achieving higher frame rates for real-time applications while maintaining high performance.

To further analyze the effectiveness of the extracted features, we conducted an additional experiment where all model parameters, along with the BatchNorm statistical parameters,  were updated during test-time adaptation. Specifically, we optimized the skeletal loss functions $\mathcal{L}_{p2s}$, $\mathcal{L}_{\mathit{sampling}}$, and $\mathcal{L}_{\mathit{radius}}$ (as described in \cref{training}) to adapt the skeletal representation to the target distribution. 

%We compared the results with MATE and demonstrated that our method outperforms it significantly, highlighting the robustness and adaptability of skeletal-based features in 3D point clouds.

\mypar{Standard Adaptation.}
In the standard mode, the model is reset after every batch, meaning each test batch adapts independently without accumulating information from previous batches. In this case, updating only the BatchNorm statistics has limited effectiveness since the adaptation does not retain context across batches. Consequently, for standard adaptation, we update all the parameters of the model using the skeletal loss functions $\mathcal{L}_{p2s}$, $\mathcal{L}_{\mathit{sampling}}$, and $\mathcal{L}_{\mathit{radius}}$. This allows the model to adjust more comprehensively to each batch of corrupted data.

\begin{table*}[t]
\setlength\tabcolsep{5pt}
\centering
\caption{Top-1 Classification Accuracy (\%) for the ModelNet-40C, ScanObjectNN-C and ShapeNet-C datasets. Differences between SMART-PC variants and MATE counterparts are indicated as {\tiny$(\pm\,\text{X}\%)$}. \textsuperscript{*} denotes reproduced results, \textsuperscript{\textdagger} denotes adaptation without backpropagation, and \textsuperscript{\ddag} denotes diffusion-based test-time adaptation methods.}
\label{tab:all-results} 
\small
% \resizebox{.62\linewidth}{!}{
\begin{tabular}{lccc}
%{lccccccccccccccc|cc}
\toprule 
& ModelNet-40C & ScanObjectNN-C & ShapeNet-C  \\
 %\multicolumn{1}{r}{Corruptions:~} & \rotb{uni} & \rotb{gauss} & \rotb{backg} & \rotb{impul} & \rotb{upsam} & \rotb{rbf} & \rotb{rbf-inv} & \rotb{den-dec} & \rotb{dens-inc} & \rotb{shear} & \rotb{rot} & \rotb{cut} & \rotb{distort} & \rotb{oclsion} & \rotb{lidar} & Avg. \\

\midrule
% Source-Only & {66.6} & {59.2} & {7.2} & {31.7} & {74.6} & {67.7} & {69.7} & {59.3} & {75.1} & {74.4} & {38.1} & {53.7} & {70.0} & {38.6} & {23.4} & {53.9} \\

 Org-SO* & 54.0 & 37.0 & 61.3 \\

 MATE-SO* & 54.4 & 34.5 & 56.5 \\
\colrow
 SMART-PC-SO & \textbf{61.7\improv{7.3}} & \textbf{38.7\improv{4.2}} & \textbf{64.5\improv{8.0}} \\
\midrule
 DUA \cite{mirza2022norm} & {54.7} & 38.7 & 60.8 \\
 TTT-Rot \cite{sun2020test} & {53.0} & -- & 60.9 \\
 SHOT \cite{liang2020we} &  {26.6} & 29.6 & 36.2 \\
 T3A \cite{iwasawa2021test} &  {55.7} & -- & 54.4 \\
 TENT \cite{wang2020tent} &  {26.5} & 27.7 & 37.4 \\
 \midrule            
 CloudFixer-Standard\textsuperscript{‡} \cite{shim2024cloudfixer} &  {68.0} & - & - \\
 CloudFixer-Online\textsuperscript{‡} \cite{shim2024cloudfixer} &  {77.2} & - & - \\
 DDA-Standard\textsuperscript{‡} \cite{gao2023back} &  {68.1} & - & - \\
 \midrule
SVWA-Standard* \cite{bahri2024test} & 57.1 & 37.4 & 50.5 \\
 MATE-Standard* \cite{mirza2023mate} & 63.0 & 36.9 & 63.1 \\
\colrow
 SMART-PC-Standard & \textbf{63.1\improv{0.1}} & \textbf{39.6\improv{2.7}} & \textbf{64.4\improv{1.3}} \\
 \midrule
 BFTT3D* \cite{wang2024backpropagation} & 57.2 & 33.0 & 60.7\\
 MATE-Online* \cite{mirza2023mate} & 70.6 & 36.9 & \textbf{69.1}\\

\colrow
 SMART-PC-Online$\dagger$ & 70.8  & 46.7 & 65.9 \\
\colrow
 SMART-PC-Online & \textbf{72.9\improv{2.3}} & \textbf{47.4\improv{10.5}} & 67.1\decr{2.0} \\
\bottomrule
\end{tabular}
% }
\end{table*}

\section{Experiments}

\begin{figure}[h]
	\centering
\includegraphics[width=1.0\linewidth]{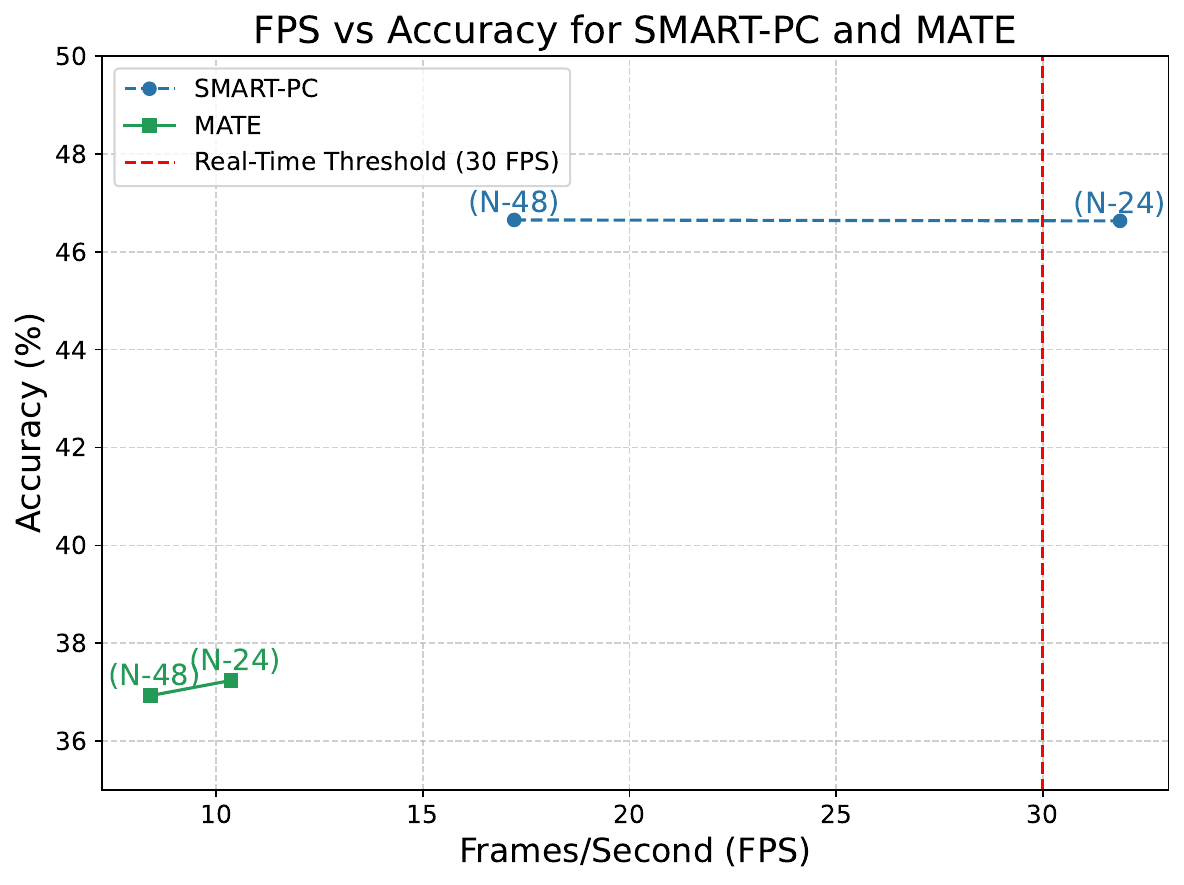}

% \vspace{-10pt}
	\caption{Frames/Second vs. Accuracy for SMART-PC and MATE during online adaptation on the ScanObjectNN dataset. SMART-PC achieves higher frame rates and accuracy, surpassing the real-time threshold (30 Frames/Second) in the (N-24) setting.}
	\label{fig:FPS}
    %\vspace{-15pt}
\end{figure}

In this section, we present a comprehensive evaluation of our proposed method using multiple 3D point cloud datasets. To thoroughly examine its robustness and generalization capabilities, we conduct experiments on three benchmark datasets: ModelNet40-C \cite{sun2022benchmarking}, ShapeNet-C \cite{chang2015shapenet}, and ScanObjectNN-C \cite{uy2019revisiting}. These datasets encompass a variety of real-world challenges, including different levels of corruption and noise, enabling us to showcase the effectiveness of our approach in diverse and complex scenarios. Furthermore, we evaluate the performance of our method in both online and standard modes. For the online mode, we compare the backpropagation-free strategy with backpropagation-based adaptation to highlight the efficiency and robustness of our approach. A detailed dataset description, additional experiments, more visualization results related to skeletons, and a detailed evaluation of our method on all datasets, including the accuracy for each corruption type, will be provided in the Supplementary Materials.

\subsection{Implementation Details}

% For both training and adaptation, we employed the Point-MAE backbone, adhering to the same settings and hyperparameters as those used in the MATE paper \cite{mirza2023mate} to ensure a fair and consistent comparison. Additionally, we evaluate the time and memory overhead associated with our method in the online mode, highlighting its computational efficiency and practicality in real-world scenarios.
For both training and adaptation, we employed the Point-MAE backbone, following the settings outlined in the MATE paper \cite{mirza2023mate} to ensure a fair and consistent comparison. The batch size is set to 1 for both adaptation modes, with 1 iteration for the online mode and 20 iterations for the standard mode, identical to the MATE paper for fair comparison once again. The optimizer and learning rate are identical to those used in the MATE paper. For augmentation, \textit{scale-transfer} is used during pre-training, similar to MATE. During adaptation, however, MATE employs random masking, whereas we use random rotation in all experiments. All experiments were conducted using a single NVIDIA A6000 GPU, ensuring consistency across all tested configurations.
Additionally, we evaluate the frames per second rate of our method and compare it with MATE.
% \mypar{Memory Overhead}

% \mypar{Time Overhead}

\mypar{Real-Time Adaptation.} \cref{fig:FPS} illustrates the adaptation performance of SMART-PC and MATE in terms of Frames/Second on the ScanObjectNN-C dataset. The adaptation strategy employed is online, with a batch size of 1 and a single iteration per adaptation step. The notation ``N-48'' and ``N-24'' represent the number of created batches, which are generated differently for the two methods: MATE uses random masking, while SMART-PC employs random rotations.

A significant advantage of SMART-PC is its ability to perform adaptation in online mode without requiring backpropagation. By updating only the statistics of BatchNorm layers, SMART-PC achieves significantly higher Frames/Second compared to MATE, which relies on computationally expensive backpropagation during adaptation.
As shown in \cref{fig:FPS}, reducing the number of batches from N-48 to N-24 does not affect the accuracy of either method. However, MATE cannot increase its Frames/Second rate due to the inherent limitations of backpropagation. In contrast, SMART-PC demonstrates a substantial increase in Frames/Second, surpassing the real-time threshold of 30 Frames/Second when fewer batches are used. This highlights the efficiency and scalability of SMART-PC, making it a practical solution for real-world applications requiring real-time adaptation. For additional comparisons with other methods, including diffusion-based approaches, please refer to the supplementary material.

% \subsection{Datasets}

% \mypar{ModelNet-40C.} ModelNet-40C \cite{sun2022benchmarking} serves as a robustness benchmark for point cloud classification, designed to evaluate the ability of architectures to handle real-world distribution shifts. It extends the original ModelNet-40 test set by introducing 15 common corruption types, grouped into three categories: transformations, noise, and density variations. These corruptions simulate practical challenges like sensor errors and LiDAR noise, offering a realistic assessment of model performance under diverse and challenging conditions.

% \mypar{ShapeNet-40C.} ShapeNetCore-v2 \cite{chang2015shapenet} is a widely used dataset for point cloud classification, containing 51,127 3D shapes spanning 55 categories. It is partitioned into three subsets: 70\% for training, 10\% for validation, and 20\% for testing. To evaluate the robustness of models under real-world conditions, \cite{mirza2023mate} augmented the test set with 15 types of corruptions, mirroring those in ModelNet-40C. These corruptions, created using the open-source implementation from \cite{sun2022benchmarking}, resulted in a modified version of the dataset known as ShapeNet-C.

\begin{figure*}[t]
    \centering
    % First figure in the left column
    \begin{minipage}[t]{0.48\textwidth}
        \centering
        \includegraphics[width=\linewidth]{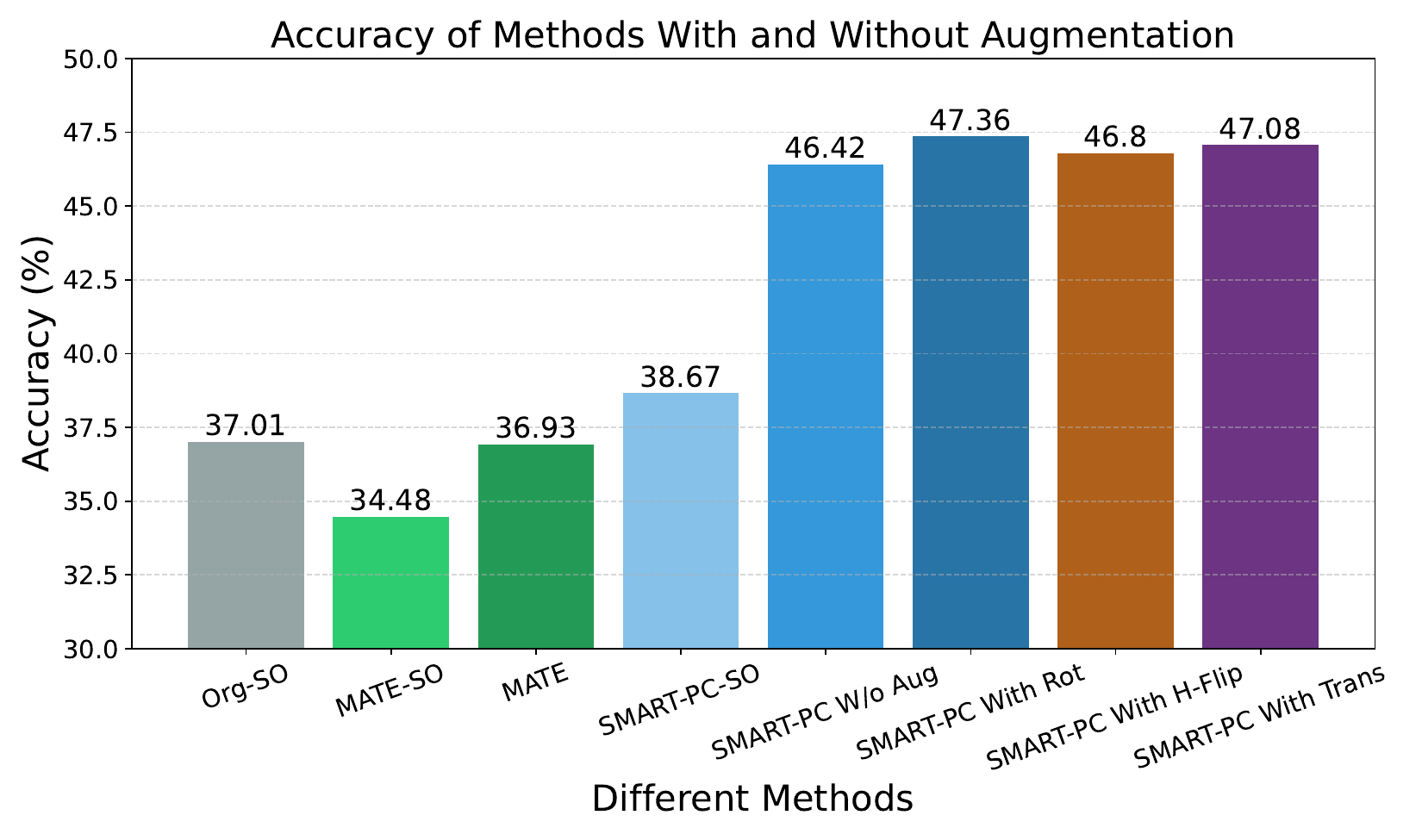}

        \vspace{-10pt}
        \caption{Effect of Augmentation During Adaptation in Our Method and Comparison with Other Methods on the ScanObjectNN Dataset in Online Mode.}
        \label{fig:different_aug}
    \end{minipage}
    \hfill
    % Second figure in the right column
    \begin{minipage}[t]{0.50\textwidth}
        \centering
        \includegraphics[width=\linewidth]{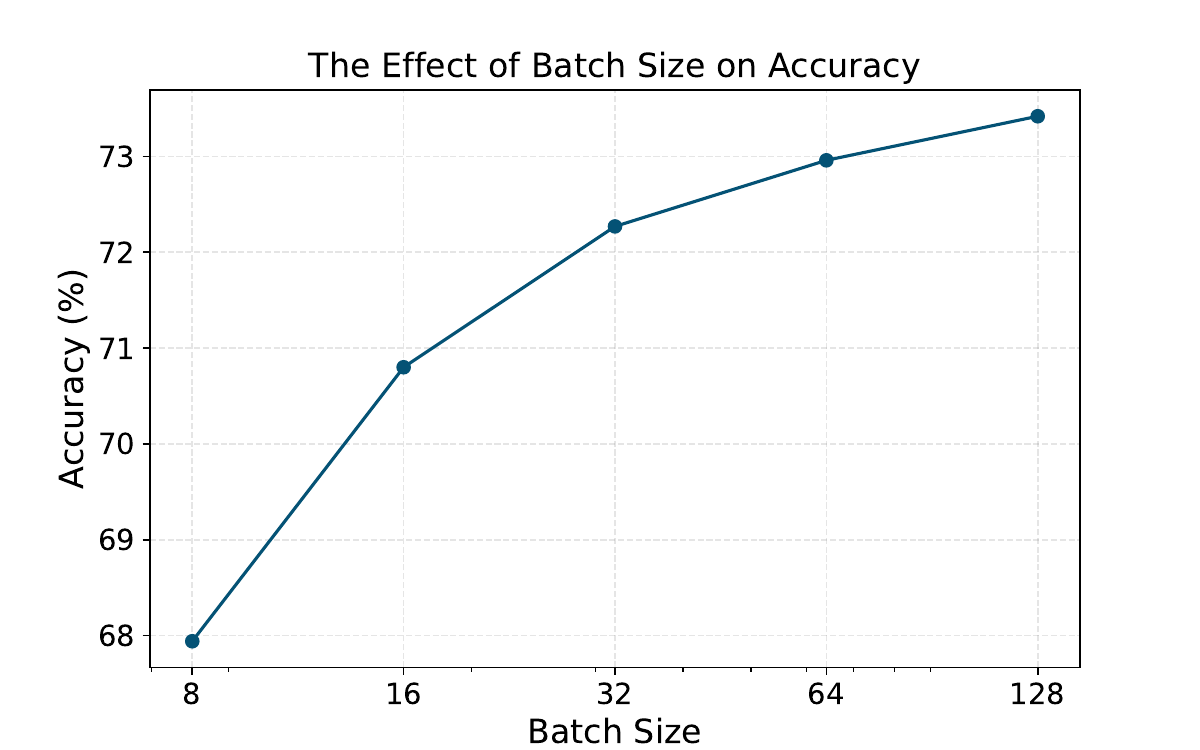}

        \vspace{-10pt}
        \caption{Impact of Batch Size on Mean Accuracy for the ModelNet-C Dataset in Standard Mode.}
        \label{fig:batch_ablation}
    \end{minipage}
\end{figure*}

% \mypar{ScanObjectNN-40C.} ScanObjectNN \cite{uy2019revisiting} is a real-world dataset for point cloud classification, comprising 2,309 training samples and 581 testing samples across 15 categories. To assess robustness, \cite{mirza2023mate} applied 15 unique corruption types to the test set, using the approach described in \cite{sun2022benchmarking}. The resulting modified dataset is referred to as ScanObjectNN-C.

% \begin{figure}
% 	\centering
% \includegraphics[width=\linewidth]{figures/Different_Augmentation_Methods.pdf}

% \vspace{-10pt}
% 	\caption{The augmentation effect on SMART-PC, tested on the ScanObjectNN dataset.}
% 	\label{fig:table_1}
%     %\vspace{-15pt}
% \end{figure}

\subsection{Main Results}

In all results tables, ``Org-SO'' refers to the evaluation of a naive pretrained model, without any additional pre-training branch, on the corrupted dataset without adaptation. ``MATE-SO'' represents a model pretrained with a reconstruction branch, evaluated on the corrupted dataset without adaptation. ``SMART-PC-SO'' denotes our model pretrained with a skeleton branch on clean data, evaluated on the corrupted dataset without adaptation. Results marked with * indicate reproduced outcomes, and ``$\dagger$'' refers to adaptation using BatchNorm statistical parameters without backpropagation.

\mypar{ModelNet-40C.} \Cref{tab:all-results} presents the top-1 classification accuracy on the ModelNet40-C dataset under various corruption types. In the source-only setting, SMART-PC-SO achieves an average accuracy of 61.7\%, significantly outperforming both MATE-SO (53.7\%) and Org-SO (54.0\%). This demonstrates the robustness of our skeleton-based pre-training in handling distribution shifts without adaptation.

For adaptation strategies, in the standard mode, SMART-PC-Standard achieves an average accuracy of 63.1\%, higher than MATE-Standard (63.0\%). In the online mode, SMART-PC-Online attains the highest average accuracy of 72.9\%, compared to 69.6\% for MATE-Online. Notably, SMART-PC-Online$\dagger$  leverages a backpropagation-free strategy, which enables efficient adaptation while maintaining superior performance. These results highlight the effectiveness and scalability of SMART-PC in both source-only (without adaptation) and adaptive settings under challenging conditions.

% Across most corruption types, SMART-PC consistently outperforms MATE, with particularly strong results in scenarios such as "impulse," "background," and "gaussian". These results highlight the effectiveness and scalability of SMART-PC in both source-only (without adaptation) and adaptive settings under challenging conditions.

\mypar{ShapeNet-C.} \Cref{tab:all-results} summarizes the top-1 classification accuracy on the ShapeNet-C dataset under various corruption types. In the source-only setting, SMART-PC-SO achieves an average accuracy of 64.5\%, outperforming both MATE-SO (56.5\%) and Org-SO (61.3\%). This result highlights the effectiveness of our skeleton-based pre-training in generalizing to unseen corruptions without adaptation.

% For adaptation strategies, in the standard mode, SMART-PC-Standard achieves an average accuracy of 64.4\%, slightly higher than MATE-Standard (63.1\%). SMART-PC-Online attains an average accuracy of 67.1\%, which is slightly lower than MATE-Online (69.1\%). This discrepancy can be attributed to MATE's reconstruction branch, which may provide an advantage in certain corruptions, such as "lidar" and "occlusion," where spatial information is heavily disrupted.
% Despite this, SMART-PC remains highly competitive and demonstrates consistent performance across most corruptions.

\mypar{ScanObjectNN-C.} \Cref{tab:all-results} presents the top-1 classification accuracy across 15 corruption types on the ScanObjectNN-C dataset. In the source-only setting, SMART-PC-SO achieves an average accuracy of 38.7\%, significantly outperforming both MATE-SO (34.5\%) and Org-SO (37.0\%). 
% This demonstrates the robustness of our skeleton-based pre-training in handling challenging corruptions without adaptation.

In the standard adaptation mode, SMART-PC-Standard achieves an average accuracy of 39.6\%, slightly higher than MATE-Standard (36.9\%). This improvement highlights the effectiveness of leveraging skeletal representations for enhancing robustness to distribution shifts.

In the online mode, SMART-PC-Online achieves the highest average accuracy of 47.4\%, significantly outperforming MATE-Online (36.9\%). Furthermore, SMART-PC$\dagger$, which utilizes a backpropagation-free adaptation strategy by updating only the BatchNorm statistics, achieves a competitive accuracy of 46.7\%. 
% SMART-PC-Online demonstrates consistent superiority across most corruption types, especially in challenging scenarios such as ``impulse noise,'' ``shear,'' ``upsample'', and ``rbf''. 
The efficiency of the backpropagation-free strategy highlights its practicality for real-world applications requiring high Frames per Second.

Overall, these results demonstrate that SMART-PC achieves state-of-the-art performance across all adaptation modes, with notable gains in real-world corruptions. 

% The method's robustness and efficiency make it a compelling solution for 3D point cloud classification under challenging conditions.

% While SMART-PC shows consistent superiority across most corruption types, it performs slightly lower than MATE in a few specific cases, such as "cut" and "occlusion." This difference can be attributed to MATE's reliance on the reconstruction branch, which may provide an advantage in handling corruptions that heavily disrupt spatial continuity. Despite this, the overall performance of SMART-PC remains highly competitive, especially considering its backpropagation-free adaptation in the online setting.

% These results further demonstrate the robustness and scalability of SMART-PC, showcasing its effectiveness in both source-only and adaptive settings on a challenging dataset like ShapeNet-C.

\subsection{Ablation Study}

% \begin{table}[!b]
% \centering
% \caption{Effect of Summation on SMART-PC Accuracy Using the ModelNet-C Dataset.}
% \label{tab:effect_add}
% \resizebox{.8\linewidth}{!}{
% \begin{tabular}{lccc}
% \toprule
% \textbf{Method} & \textbf{With Sum.} & \textbf{Without Sum.} & \textbf{Acc. (\%)} \\ 
% \midrule
% SMART-PC & \(\times\) & \checkmark & 62.4 \\ 
% SMART-PC & \checkmark & \(\times\) & \textbf{63.1} \\ 
% \bottomrule
% \end{tabular}
% }
% \end{table}

\mypar{Batch Size.}
We evaluate the effect of batch size on our method's performance using the ModelNet40-C dataset in standard adaptation mode, where all parameters are updated, with 20 iterations as in the MATE paper. As shown in \cref{fig:batch_ablation}, increasing the batch size improves accuracy, from 67.94\% at batch size 8 to 73.42\% at batch size 128. This improvement showcases the effectiveness of larger batch sizes for achieving higher accuracy. To ensure fairness in comparison with MATE and other methods and to avoid increasing computational costs, we used a batch size of 1 for all main results in both standard and online modes.

% \mypar{Number of Iterations.} blo blo

\mypar{Augmentation.} 
\Cref{fig:different_aug} illustrates the impact of different augmentations and the absence of augmentation on our method during adaptation. This experiment was conducted on the ScanObjectNN dataset in online mode with batch size 1 and iteration 1. Consistent with the MATE paper, we created 48 batches using random rotations to introduce diversity among the data. Even without augmentation (\textit{SMART-PC W/O Aug}), our method achieves 46.42\% accuracy, outperforming \textit{MATE} (36.93\%), demonstrating the robustness of our skeleton-based framework.
Applying random rotations (\textit{SMART-PC With Rot}) increases accuracy to 47.36\%. Other augmentations, including horizontal flipping (\textit{SMART-PC With H-Flip}) and translations (\textit{SMART-PC With Trans}), achieve comparable improvements, with accuracies of 46.8\% and 47.08\%, respectively.

% \begin{figure}[t]
% 	\centering
% \includegraphics[width=\linewidth]{figures/Augmentation_Methods_2.pdf}

% % \vspace{-12pt}
% 	\caption{Effect of Augmentation During Adaptation in Our Method and Comparison with Other Methods on the ScanObjectNN Dataset in Online Mode}
% 	\label{fig:different_aug}
%     %\vspace{-15pt}
% \end{figure}

% \begin{figure}[t!]
% 	\centering
% \includegraphics[width=.85\linewidth]{figures/ablation_batch.pdf}

% % \vspace{-10pt}
% 	\caption{Impact of Batch Size on Mean Accuracy for the ModelNet-C Dataset in Standard Mode}
% 	\label{fig:batch_ablation}
%     %\vspace{-15pt}
% \end{figure}

% \CD{You should also consider measuring the FLOPs runtime to assess the claim that our method is more efficient computationally...}
\mypar{Feature Summation.} 
We conduct an ablation study on the impact of summing the features from the shared encoder, used for both skeletal and classification tasks, with the decoder features. This summation improves SMART-PC’s accuracy from 62.4\% to 63.1\% on the ModelNet-C dataset in standard mode, compared to using only the encoder features for classification.

% \CD{Are these actual results? If so, why not test larger batch size, at least until the accuracy plateaus?}

\begin{table*}[h]
\centering
\small
\begin{tabular}{ccccc}
\toprule
Pt2Sphere & {Sampling} & {RadiusReg} & {Source Acc(\%)} & {Corrupted Acc(\%)} \\
\midrule
1.0 & 1.0 & 0.0 & 91.3 & 67.82 \\
0.0 & 1.0 & 1.0 & 91.6 & 67.79 \\
1.0 & 0.0 & 1.0 & 91.6 & 67.80 \\
1.0 & 1.0 & 1.0 & 91.2 & 72.84 \\
\textbf{0.3} & \textbf{1.0} & \textbf{0.4} & \textbf{91.3} & \textbf{72.95} \\
\bottomrule
\end{tabular}
\caption{Ablation study of skeleton loss coefficients on ModelNet40 and ModelNet40-C (online adaptation). The best configuration corresponds to the original coefficients from the Point2Skeleton paper.}
\label{tab:skeleton_ablation}
\end{table*}

\begin{table*}[h]
\centering
\small
\begin{tabular}{lccc}
\toprule
{Dataset} & {Org-SO} & {MATE-SO} & {SMART-PC-SO} \\
\midrule
ScanObjectNN-C & 33.00 & 33.22 & \textbf{35.90} \\
ModelNet40-C   & 57.16 & 54.71 & \textbf{65.25} \\
ShapeNet-C     & 60.73 & 53.07 & \textbf{62.24} \\
\bottomrule
\end{tabular}
\caption{Mean accuracy (\%) of BFTT3D using different pretrained models under the backpropagation-free setting. SMART-PC-SO achieves the best results across all datasets.}
\label{tab:pretrain_ablation}
\end{table*}

\mypar{Skeleton Loss Coefficients.} 
To further evaluate the contribution of each loss component in our skeleton-based pretraining, we conducted an ablation study using different coefficients for the Skeletal loss terms: point-to-sphere loss, sampling loss, and radius regularization. The experiments were performed on the ModelNet40 and ModelNet40-C datasets under the online adaptation setting.

As shown in ~\cref{tab:skeleton_ablation}, the best performance is obtained with the coefficient set $(0.3, 1.0, 0.4)$, which corresponds to the original settings in the Point2Skeleton paper \cite{lin2021point2skeleton}. This configuration achieves the highest corrupted accuracy of 72.95\%, confirming that each skeletal loss term contributes meaningfully to the learning of robust features under corruption. Additionally, as described in the main paper, the Radius Regularization Loss (Equation 6) is designed to avoid instability caused by overly small radii, especially under noisy conditions. This loss encourages the model to learn larger and more stable radii, which improves the robustness of the skeletal abstraction. Although we do not observe excessively large radii, the Point-to-Sphere and Sampling losses (Equations 11 and 12) implicitly constrain radius size by preserving geometric consistency. As shown in ~\cref{tab:skeleton_ablation}, removing the regularization term leads to a drop in performance, confirming its importance.

\mypar{Evaluating Pretraining Strategies in Backpropagation-Free Adaptation.} 
Our method supports two adaptation modes: one with backpropagation and one that is backpropagation-free. The goal of the backpropagation-free mode is to show that pretraining with a skeleton-based decoder encourages the model to learn robust and meaningful geometric features. These features are resilient enough that, during test-time, simply updating the BatchNorm statistical parameters (i.e., running mean and variance) is sufficient to improve performance—without performing gradient-based updates.

To validate this effect, we conducted additional experiments using the BFTT3D~\cite{wang2024backpropagation} adaptation method across three different pretraining strategies (Org-SO, MATE-SO, and SMART-PC-SO).
% \begin{itemize}
%     \item \textbf{Org-SO:} A baseline model pretrained with only an encoder and classification head.
%     \item \textbf{MATE-SO:} A model pretrained using the MATE framework, which includes an encoder, a reconstruction decoder, and a classification head.
%     \item \textbf{SMART-PC-SO:} Our model, pretrained with an encoder, a skeleton-based decoder, and a classification head.
% \end{itemize}
Each pretrained model was evaluated using the same BFTT3D adaptation strategy under the backpropagation-free setting. As shown in Table~\ref{tab:pretrain_ablation}, our SMART-PC-SO model consistently outperforms both Org-SO and MATE-SO across all three datasets. This provides strong evidence that the skeletal decoder encourages the model to extract more structure-aware and corruption-resilient features, which support effective test-time adaptation without updating model weights.

\section{Conclusion}

% In this paper, we proposed SMART-PC, a skeleton-based framework for robust and efficient test-time training of 3D point cloud models under distribution shifts. By leveraging skeletal representations, our method enables the model to extract meaningful geometric features, which enhance robustness to corruptions. Unlike previous approaches, SMART-PC performs online adaptation with a backpropagation-free strategy by updating only BatchNorm statistics, making it computationally efficient. 
% Extensive experiments on multiple benchmark datasets, including ModelNet40-C, ShapeNet-C, and ScanObjectNN-C, demonstrated that SMART-PC achieves state-of-the-art performance in both source-only (without adaptation) and adaptation settings.
% Overall, SMART-PC provides a scalable and practical solution for real-world applications requiring robust 3D point cloud classification under challenging conditions. Future work may explore extending this framework to other 3D tasks, such as segmentation and object detection.

In this paper, we proposed SMART-PC, a skeleton-based framework for robust and efficient test-time training of 3D point cloud models under distribution shifts. By leveraging skeletal representations, SMART-PC enhances robustness to corruptions while enabling high Frames per Second during online adaptation through a backpropagation-free strategy that updates only BatchNorm statistics. Experiments on benchmark datasets, including ModelNet40-C, ShapeNet-C, and ScanObjectNN-C, demonstrate state-of-the-art performance in both source-only and adaptation settings. Overall, SMART-PC provides a scalable and practical solution for real-world applications requiring robust 3D point cloud classification under challenging conditions. Future work may explore extending this framework to other 3D tasks, such as semantic segmentation, instance segmentation, and object detection.

%% file: sec/supp.tex
\clearpage
\setcounter{page}{1}
\setcounter{section}{0}
\renewcommand{\thesection}{\Alph{section}}
\renewcommand{\thesubsection}{\thesection.\arabic{subsection}}
\maketitlesupplementary

\section{Implementation}
Our approach was implemented using PyTorch, with the codebase organized into two main components: \emph{Pretrain} and \emph{Adaptation}.

\paragraph{Pretrain.}
We start with the initial pretraining phase of the base models (Point-MAE). In this phase, the backbone is pretrained with a classification branch in a fully supervised manner and a skeleton branch in a self-supervised manner. The pretraining is performed on clean datasets such as ModelNet, ShapeNet, and ScanObjectNN, ensuring the models are well-prepared for the subsequent adaptation steps.

\paragraph{Adaptation.}
After completing the pretraining phase, we transition to the adaptation stage, which consists of two modes: \textbf{online} and \textbf{standard}.
In the \textbf{standard adaptation mode}, all model parameters are updated using the skeleton loss, allowing the model to comprehensively adjust to the target data.
In the \textbf{online adaptation mode}, we employ two distinct strategies. The first strategy involves adapting only the statistical parameters of the BatchNorm layers (e.g., running mean and variance) without backpropagation. This approach significantly reduces computational costs, enabling our method to achieve higher Frames per Second and making it suitable for real-time applications. The second strategy involves updating all model parameters using the skeleton loss, similar to the standard mode.
These flexible adaptation strategies highlight the efficiency and scalability of our method, catering to both real-time and high-accuracy requirements.
% The code will be published upon the acceptance of the paper.

\section{Datasets}

\mypar{ModelNet-40C.} ModelNet-40C \cite{sun2022benchmarking} serves as a robustness benchmark for point cloud classification, designed to evaluate the ability of architectures to handle real-world distribution shifts. It extends the original ModelNet-40 test set by introducing 15 common corruption types, grouped into three categories: transformations, noise, and density variations. These corruptions simulate practical challenges like sensor errors and LiDAR noise, offering a realistic assessment of model performance under diverse and challenging conditions.

\mypar{ShapeNet-C.} ShapeNetCore-v2 \cite{chang2015shapenet} is a widely used dataset for point cloud classification, containing 51,127 3D shapes spanning 55 categories. It is partitioned into three subsets: 70\% for training, 10\% for validation, and 20\% for testing. To evaluate the robustness of models under real-world conditions, \cite{mirza2023mate} augmented the test set with 15 types of corruptions, mirroring those in ModelNet-40C. These corruptions, created using the open-source implementation from \cite{sun2022benchmarking}, resulted in a modified version of the dataset known as ShapeNet-C.

\mypar{ScanObjectNN-C.} ScanObjectNN \cite{uy2019revisiting} is a real-world dataset for point cloud classification, comprising 2,309 training samples and 581 testing samples across 15 categories. To assess robustness, \cite{mirza2023mate} applied 15 unique corruption types to the test set, using the approach described in \cite{sun2022benchmarking}. The resulting modified dataset is referred to as ScanObjectNN-C.

% \begin{table*}
% \centering
% \small
% \begin{tabular}{ccccc}
% \toprule
% \textbf{Pt2Sphere} & \textbf{Sampling} & \textbf{RadiusReg} & \textbf{Source Acc(\%)} & \textbf{Corrupted Acc(\%)} \\
% \midrule
% 1.0 & 1.0 & 0.0 & 91.3 & 67.82 \\
% 0.0 & 1.0 & 1.0 & 91.6 & 67.79 \\
% 1.0 & 0.0 & 1.0 & 91.6 & 67.80 \\
% 1.0 & 1.0 & 1.0 & 91.2 & 72.84 \\
% \textbf{0.3} & \textbf{1.0} & \textbf{0.4} & \textbf{91.3} & \textbf{72.95} \\
% \bottomrule
% \end{tabular}
% \caption{Ablation study of skeleton loss coefficients on ModelNet40 and ModelNet40-C (online adaptation). The best configuration corresponds to the original coefficients from the Point2Skeleton paper.}
% \label{tab:skeleton_ablation}
% \end{table*}

\section{More Experiments}

\mypar{Comparison and Advantages of SMART-PC over BFTT3D in Backpropagation-Free Mode.} Unlike BFTT3D, which uses class-based prototypes from the source dataset during adaptation, our method relies solely on the target dataset without accessing any source information. Another key distinction is efficiency: as shown in \cref{tab:fps_comparison}, our method in the backpropagation-free mode achieves significantly higher FPS compared to MATE, SVWA, BFTT3D, CloudFixer, and DDA. In terms of performance, our method also outperforms BFTT3D across three corrupted datasets, as reported in \cref{tab:all-results}. Together, \cref{tab:fps_comparison} and \cref{tab:all-results} demonstrate that our skeleton-based pretraining enables the model to learn robust geometric features, allowing effective adaptation with only statistical parameter updates, achieving both high accuracy and fast inference. 

\begin{table}
\centering
\small
\begin{tabular}{lc}
\toprule
\textbf{Method} & \textbf{FPS} \\
\midrule
DDA & 0.04 \\
CloudFixer & 1.07 \\
BFTT3D & 6.83 \\
SVWA & 10.86 (for $N_v{=}2$) \\
MATE & 10.79 \\
\textbf{SMART-PC} & \textbf{59.52} \\
\bottomrule
\end{tabular}
\caption{Comparison of inference speed (FPS) across different adaptation methods on the ModelNet40-C dataset.}
\label{tab:fps_comparison}
\end{table}

\mypar{Analyzing the Impact of BatchNorm Statistic Updates.} 
Updating the running mean and variance in BatchNorm layers has been shown to improve robustness under covariate shift \cite{nado2020evaluating}. Furthermore, the ADABN \cite{li2018adaptive} paper supports the significance of this mechanism by stating that " label related knowledge is stored in the weight matrix of each layer, whereas domain related knowledge is represented by the statistics of the Batch Normalization (BN)" This highlights that updating BN statistics is a meaningful and effective approach for handling domain shifts.

During pretraining, our method learns more abstract and robust features through the skeleton prediction branch. These features are resilient to corruption, such that during test-time adaptation, simply updating the statistical parameters of the BatchNorm layers (i.e., running mean and variance) can effectively suppress noise without requiring backpropagation. 

\begin{figure*}
\centering
\includegraphics[width=0.6\textwidth]{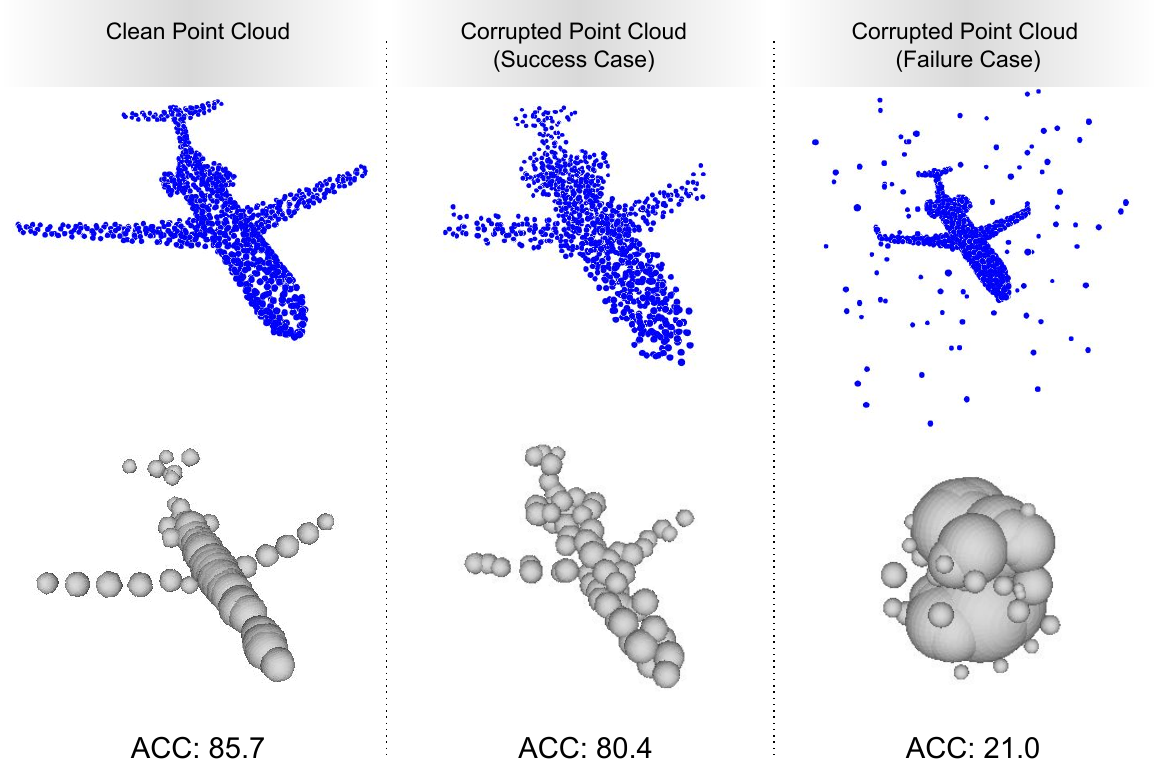}

\vspace{-12pt}
\caption{Visualization of a clean point cloud and its corrupted versions. The predicted skeleton remains stable even under corruption, supporting accurate classification (ACC: 85.7 and 80.4 in the clean and success cases, respectively). In the failure case, the corruption leads to misaligned skeletons and significantly lower performance (ACC: 21.0).
}
\label{corrupt_source}
\end{figure*}

As shown in \cref{corrupt_source}, our method maintains strong accuracy even under corruption. The clean input (left) yields a high classification accuracy of 85.7\%, while the corrupted input still achieves 80.4\% in a successful case, demonstrating robustness. In contrast, a failure case (right) leads to significant misalignment, dropping accuracy to 21.0\%. These results indicate that the learned skeletal representation acts as a noise-resistant abstraction, preserving semantic structure under moderate distortions.

\begin{figure*}[!t]
\centering
\includegraphics[width=1.0\textwidth]{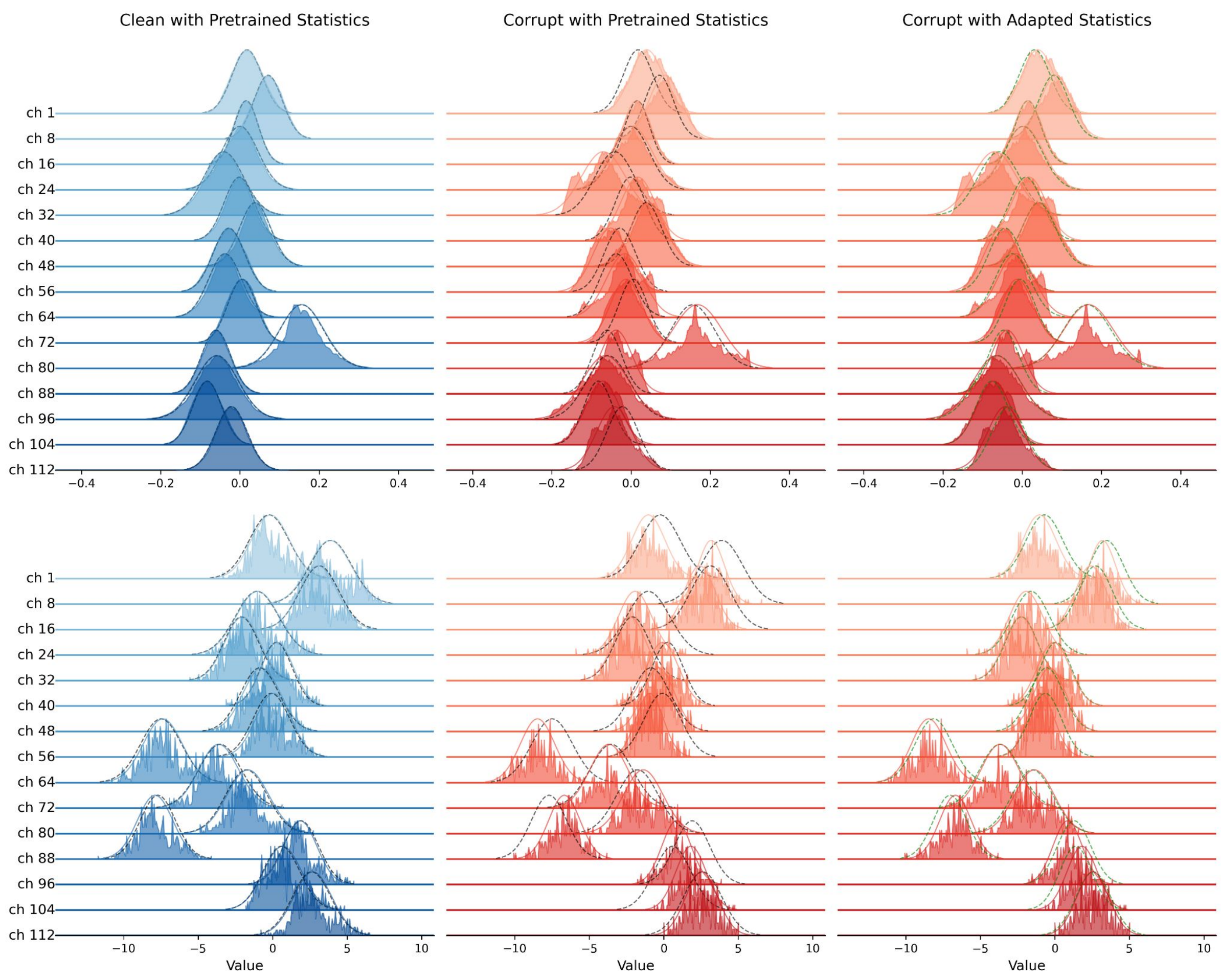}

\vspace{-12pt}
\caption{Distribution of the input (solid colors) and the running statistics (dashed curve) of BatchNorm layers across different channels. Top: first BN layer in the encoder, Bottom: first BN layer in the classifier. Blue: clean source data, Red: corrupted target data, Solid Curve: Gaussian curve of data statistics, Black Dashed Curve: pretrained accumulated
statistics, Green Dashed Curve: adapted statistics.
}
\label{bn_analyze}
\end{figure*}

To further analyze the effect of BatchNorm statistics update in our setting, we visualized the impact of distribution shift on the BatchNorm input and how updating the statistics can help realign the model to the new distribution. In \cref{bn_analyze}, the Gaussian solid curves represent the statistics of the input data. As observed, the source distribution (blue) aligns well with the accumulated statistics from pre-training (black dashed curve), resulting in a centered and scaled distribution with zero mean and unit variance. However, when facing a distribution shift (red) in the center column, the pre-training-time accumulated statistics no longer align with the corrupted target distribution, leading to an inconsistent input distribution to the subsequent layers—compared to what the model has seen during training (i.e., covariate shift). This misalignment is clearly visible in the center column as the distance between the red solid curve and the black dashed curve. By updating the BatchNorm statistics, the running mean and variance shift toward the target distribution. This helps mitigate the covariate shift introduced by the target domain (right column). It is evident that updating the batch statistics moves the BatchNorm statistics (green dashed curves) closer to the data distribution (solid curves).

This pattern is consistent across different channels of both the first BatchNorm layer in the encoder (top row) and the classification head (bottom row). For each BN layer, channel input values are aggregated across batch samples and tokens, and plotted as histograms of the values. Similar to the input data statistics, the BN statistics curves are plotted as Gaussian curves, computed from the corresponding running mean and running variance.

\section{Detailed Results}

In this section, the performance of our method is presented in \cref{tab:modelnet-c-results}, \cref{tab:shapenet-c-results}, and \cref{tab:scanobject-c-results} showcasing detailed results for each corruption across all datasets. Our method consistently outperforms previous approaches in source-only (without adaptation), standard adaptation, and online adaptation modes. In most corruption scenarios, SMART-PC achieves higher accuracy compared to prior methods, demonstrating its robustness and effectiveness in handling distribution shifts across ModelNet40-C, ShapeNet-C, and ScanObjectNN-C. The improvements are particularly significant in challenging corruption types, highlighting the advantages of leveraging skeletal representations for test-time training.

\begin{table*}[t]
\setlength\tabcolsep{4.5pt}
\centering
    \small
\caption{Top-1 Classification Accuracy (\%) for all distribution shifts in the ModelNet40-C dataset. All results are based on the PointMAE backbone trained on a clean training set and adapted to the OOD test set with a batch size of 1.
Results marked with * indicate reproduced outcomes (adaptation only) using the pretrained models from the MATE GitHub repository. Results marked with ** indicate full reproduction from scratch, including both pretraining and adaptation. The symbol ``$\dagger$'' denotes adaptation using only BatchNorm statistical parameter updates, without backpropagation.}
\label{tab:modelnet-c-results}    
\resizebox{.9\linewidth}{!}{
\begin{tabular}{llllllllllllllll|l}
%{lccccccccccccccc|cc}
\toprule
      \multicolumn{1}{r}{Corruptions:~}           &  \rotb{uni} &  \rotb{gauss} &  \rotb{backg} &  \rotb{impul} &  \rotb{upsam} &   \rotb{rbf} &   \rotb{rbf-inv}  &  \rotb{den-dec} & \rotb{dens-inc} &   \rotb{shear}  &  \rotb{rot}  &   \rotb{cut} &   \rotb{distort} &   \rotb{oclsion}  &    \rotb{lidar} &   Avg. \\

\midrule
% Source-Only &             {66.6} &             {59.2} &              {7.2} &             {31.7} &             {74.6} &             {67.7} &             {69.7} &             {59.3} &             {75.1} &             {74.4} &             {38.1} &             {53.7} &             {70.0} &             {38.6} &             {23.4} &             {53.9} \\

     Org-SO* &            66.6	&59.2	& \phz7.2	&31.8	&74.6 &	67.7 &	69.8 &	59.3 &	75.1&	74.4&	38.0&	53.7&	70.0 &	\textbf{38.6}&	23.4&	54.0 \\

    MATE-SO* & 59.7&	51.3&	\textbf{28.2}&	55.3&	71.5	&57.4&	60.7&	65.2&	\textbf{77.4}&	67.1&	30.2&	62.3&	61.9&	37.2&	19.9&	53.7 \\
    MATE-SO** & 60.8&	53.3&	28.8&	53.4&	70.9&	57.9&	59.6&	\textbf{68.9}&	77.1&	67.5&	31.7&	65.6&	62.0&	33.9&	\textbf{24.6} & 54.4 \\

\colrow
  SMART-PC-SO & \textbf{81.8}&	\textbf{79.5}&	13.6&	\textbf{65.4}&	\textbf{84.3}&	\textbf{75.7}	&\textbf{77.8}&	62.0&	65.9&	\textbf{73.4}&	\textbf{42.7}	&\textbf{69.1}	&\textbf{73.8}	&36.3&	24.4&	\textbf{61.7} \\
\midrule
           DUA &             {65.0} &             {58.5} &             {14.7} &             {48.5} &             {68.8} &             {62.8} &             {63.2} &             {62.1} &             {66.2} &             {68.8} & {46.2} &             {53.8} &             {64.7} & {{41.2}} & {{36.5}} &             {54.7} \\
           TTT-Rot &             {61.3} &             {58.3} &  {{34.5}} &             {48.9} &             {66.7} &             {63.6} &             {63.9} &             {59.8} &             {68.6} &             {55.2} &             {27.3} &             {54.6} &             {64.0} &             {40.0} &             {29.1} &             {53.0} \\
          SHOT &             {29.6} &             {28.2} &              {9.8} &             {25.4} &             {32.7} &             {30.3} &             {30.1} &             {30.9} &             {31.2} &             {32.1} &             {22.8} &             {27.3} &             {29.4} &             {20.8} &             {18.6} &             {26.6} \\
           T3A &             {64.1} &             {62.3} & {{33.4}} &             {65.0} &             {75.4} &             {63.2} &             {66.7} &             {57.4} &             {63.0} &             {72.7} &             {32.8} &             {54.4} &             {67.7} &             {39.1} &             {18.3} &             {55.7} \\
          TENT &             {29.2} &             {28.7} &             {10.1} &             {25.1} &             {33.1} &             {30.3} &             {29.1} &             {30.4} &             {31.5} &             {31.8} &             {22.7} &             {27.0} &             {28.6} &             {20.7} &             {19.0} &             {26.5} \\
        \midrule
 MATE-Standard* & 69.8&	61.8	&\textbf{18.9}&	63.9&	72.5	&64.0&	66.0&	\textbf{74.0}&	\textbf{80.8}&	71.0	&36.7&	69.2	&66.3& \textbf{38.4}	&\textbf{29.9}&	58.9 \\

 MATE-Standard** &  75.3 &	70.9 &	23.0 &	64.7	&79.2	&67.9	&69.3&	76.5 &	84.0 &	75.9 &	47.2 &	71.8&	71.6 &	37.5 &	29.7&	63.0 \\

\colrow
  SMART-PC-Standard & \textbf{82.4}&	\textbf{80.1}&	12.0	&\textbf{67.1}	&\textbf{84.5}	&\textbf{76.0}&	\textbf{78.6}	&67.3	&72.9	&\textbf{73.3}& \textbf{43.9}	&\textbf{72.6}&	\textbf{73.5}	&37.4&	24.8&	\textbf{63.1} \\
  \midrule
   MATE-Online* &  80.6	&79.5	&20.7&	71.5	&82.6&	78.1& 80.7	&78.1&	\textbf{86.6}&	\textbf{79.6}	&54.9	&78.4&	77.4&	\textbf{45.4}&	\textbf{49.6} & 69.6\\

     MATE-Online** &  82.1&	78.3 &	29.6&	74.1&	84.3&	77.4	&79.2&	79.4&	86.7	&78.8&	55.2&	79.1&	76.9&	47.8&	49.8&	70.6 \\

\colrow
    SMART-PC-Online$\dagger$ & 85.0&	83.2&	31.6&	77.6&	\textbf{85.9}&	79.2&	80.8&	77.8&	79.3&	77.8&	60.3&	80.6	&\textbf{76.8}&	45.2	&40.4&	\textbf{70.8} \\
\colrow
       SMART-PC-Online & \textbf{85.4}	&\textbf{84.0}&	\textbf{49.4}&	\textbf{79.7}&	85.7&	\textbf{80.1}&	\textbf{81.3}&	\textbf{81.7}&	82.7&	78.3&	\textbf{60.5}&	\textbf{82.6}&	76.7&	44.4&	41.8&	\textbf{72.9} \\

\bottomrule
\end{tabular}
}
\end{table*}

\begin{table*}[t]
\setlength\tabcolsep{4.5pt}
\centering
    \small
\caption{Top-1 Classification Accuracy (\%) for all distribution shifts in the ShapeNet-C dataset. All results are based on the PointMAE backbone trained on a clean training set and adapted to the OOD test set with a batch size of 1. Results marked with * indicate reproduced outcomes, while ``$\dagger$'' denotes adaptation using BatchNorm statistical parameters without backpropagation.
% \emph{Source-Only} denotes its performance on the corrupted test data without any adaptation. 
% Highest Accuracy is shown in bold, while second best is underlined.
}
\label{tab:shapenet-c-results}    
\resizebox{.9\linewidth}{!}{
\begin{tabular}{llllllllllllllll|l}
\toprule
      \multicolumn{1}{r}{Corruptions:}           &  \rotb{uni} &  \rotb{gauss} &  \rotb{backg} &  \rotb{impul} &  \rotb{upsam} &   \rotb{rbf} &   \rotb{rbf-inv}  &  \rotb{den-dec} & \rotb{dens-inc} &   \rotb{shear}  &  \rotb{rot}  &   \rotb{cut} &   \rotb{distort} &   \rotb{oclsion}  &    \rotb{lidar} &   Avg. \\

\midrule
    % Source-Only &             {69.2} &             {62.8} &             {10.3} &             {56.2} &             {70.1} &             {70.5} &             {71.9} & {\underline{85.5}} & {\underline{86.2}} &             {73.9} &             {41.3} & {\underline{84.4}} &             {69.9} &              {7.9} &              {3.9} &             {57.6} \\

    Org-SO*  &77.4 &71.8	&\phz8.6 &	54.4&	77.9&	75.5	&76.0	&85.3&	76.5&	80.5&	57.1&	85.1&	76.0&	11.0&	\phz7.1&	61.3 \\

    MATE-SO* & 69.7&	63.3&	\phz2.1&	50.6&	71.1&	70.2&	72.1&	\textbf{85.9}&	\textbf{77.8}&	75.6&	44.0&	\textbf{85.4}&	70.3&	\phz7.0&	\phz3.1&	56.5 \\
    \colrow
   SMART-PC-SO & \textbf{80.6}&	\textbf{78.5}&	\textbf{11.4}&	\textbf{61.3}&	\textbf{81.6}&	\textbf{81.1}&	\textbf{81.5}&	84.9&	74.4&	\textbf{81.1}&	\textbf{64.1}&	85.0&	\textbf{79.9}&	\textbf{11.8}&	\textbf{10.0}&	\textbf{64.5} \\

    \midrule
           DUA &             {76.1} &             {70.1} &             {14.3} &             {60.9} &             {76.2} &             {71.6} &             {72.9} &             {80.0} &             {83.8} &             {77.1} & {{57.5}} &             {75.0} &             {72.1} &             {11.9} &             {12.1} &             {60.8} \\
       TTT-Rot &             {74.6} &             {72.4} & {{23.1}} &             {59.9} &             {74.9} &             {73.8} &             {75.0} &             {81.4} &             {82.0} &             {69.2} &             {49.1} &             {79.9} &             {72.7} & {{14.0}} &             {12.0} &             {60.9} \\
          SHOT &             {44.8} &             {42.5} &             {12.1} &             {37.6} &             {45.0} &             {43.7} &             {44.2} &             {48.4} &             {49.4} &             {45.0} &             {32.6} &             {46.3} &             {39.1} &              {6.2} &              {5.9} &             {36.2} \\
           T3A &             {70.0} &             {60.5} &              {6.5} &             {40.7} &             {67.8} &             {67.2} &             {68.5} &             {79.5} &             {79.9} &             {72.7} &             {42.9} &             {79.1} &             {66.8} &              {7.7} &              {5.6} &             {54.4} \\
          TENT &             {44.5} &             {42.9} &             {12.4} &             {38.0} &             {44.6} &             {43.3} &             {44.3} &             {48.7} &             {49.4} &             {45.7} &             {34.8} &             {48.6} &             {43.0} &             {10.0} &             {10.9} &             {37.4} \\
        \midrule
 MATE-Standard & {{77.8}} & {{74.7}} &              {4.3} & {\textbf{66.2}} & {{78.6}} & {{76.3}} & {{75.3}} &  {\textbf{86.1}} &  {\textbf{86.6}} & {{79.2}} &             {56.1} &             {84.1} & {{76.1}} &             \textbf{12.3} & \textbf{{13.1}} & {{63.1}} \\
\colrow
 SMART-PC-Standard & \textbf{80.8}	&\textbf{78.9}&	 \textbf{8.9}&	60.4&	\textbf{81.8}&	\textbf{81.1}&	\textbf{81.7}&	84.8&	78.4&	\textbf{80.8}	&\textbf{63.7}&	\textbf{84.9}&	\textbf{79.8}&	11.5&	\phz8.8&	\textbf{64.4}\\

\midrule
 
   MATE-Online* &  \textbf{81.5} &  {{78.6}} &  {\textbf{40.9}} &  {\textbf{75.9}} &  {\textbf{81.6}} &  {{79.7}} &  {{80.1}} &            \textbf{84.9} &             \textbf{85.9} &  {\textbf{81.8}} &  {{70.8}} &  {\textbf{85.1}} &  {{79.0}} &  {\textbf{14.2}} &  {\textbf{16.6}} &  {\textbf{69.1}} \\

\colrow
SMART-PC-Online$\dagger$ & 80.4	&78.7&	21.0&	72.7&	80.9&	80.9&	\textbf{80.6}	&82.5	&78.3	&80.9	&70.1	&82.5	&79.0	&10.5	&\phz9.7	&65.9\\
\colrow
 SMART-PC-Online & 81.2	&\textbf{80.5}&	28.9&	74.3&	81.2	&\textbf{80.7}&	80.5&	83.1&	81.0&	80.4	&\textbf{73.2}	&82.8	&\textbf{79.0}&	10.0&	10.2&	67.1 \\
\bottomrule
\end{tabular}
}
\end{table*}

\begin{table*}[t]
\setlength\tabcolsep{4.5pt}
\centering
    \small
\caption{Top-1 Classification Accuracy (\%) for all distribution shifts in the ScanObjectNN-C dataset. All results are based on the PointMAE backbone trained on a clean training set and adapted to the OOD test set with a batch size of 1. Results marked with * indicate reproduced outcomes (adaptation only) using the pretrained models from the MATE GitHub repository. Results marked with ** indicate full reproduction from scratch, including both pretraining and adaptation. The symbol ``$\dagger$'' denotes adaptation using only BatchNorm statistical parameter updates, without backpropagation.
% \emph{Source-Only} denotes its performance on the corrupted test data without any adaptation. 
% Highest Accuracy is shown in bold, while second best is underlined.
}
\label{tab:scanobject-c-results}  
\resizebox{.9\linewidth}{!}{
\begin{tabular}{llllllllllllllll|l}
%{lccccccccccccccc|cc}
\toprule
      \multicolumn{1}{r}{Corruptions:~}           &  \rotb{uni} &  \rotb{gauss} &  \rotb{backg} &  \rotb{impul} &  \rotb{upsam} &   \rotb{rbf} &   \rotb{rbf-inv}  &  \rotb{den-dec} & \rotb{dens-inc} &   \rotb{shear}  &  \rotb{rot}  &   \rotb{cut} &   \rotb{distort} &   \rotb{oclsion}  &    \rotb{lidar} &   Avg. \\
\midrule
     Org-SO* &            21.7&	18.8	&16.9 &	18.4 &	22.2 &	\textbf{46.0} &	47.0 &	\textbf{72.1}	&\textbf{69.4}	&\textbf{48.9}	&35.6	&\textbf{73.0} &	\textbf{49.4} &	6.7 &	9.3 &	37.0 \\
     MATE-SO* &             20.3&	32.2&	\textbf{18.9}&	21.2	&20.5&	35.6&	36.7&	69.9	&66.6&	38.9&	28.7&	70.4&	39.4&	\phz8.3&	\phz9.8	& 34.5 \\
     MATE-SO** &             15.3 &	20.1 &	13.4 &	11.0 &	15.3 &	28.7 &	29.4 &	69.2 &	64.5 &	32.9 &	25.0 &	70.6	&33.4 &	9.1&	9.0 & 29.8 \\
     \colrow
    SMART-PC-SO &             \textbf{26.7}&	\textbf{37.7}&	16.9&	\textbf{21.3}&	\textbf{27.2}&	44.2&	\textbf{48.9}&	69.5&	56.3&	48.5&	\textbf{43.2}&	72.3&	48.0&	\phz\textbf{8.4}	&\textbf{11.0}&	\textbf{38.7}\\
\midrule
    
           DUA* &            30.5	&40.1	&10.2&	23.6&	29.9&	43.7	&46.1&	68.3&	66.3&	48.5	&38.9&	68.7&	48.4&	\phz8.6	&\phz8.1	&38.7 \\
       % TTT-Rot &            - &              - &              - &              - &              - &             - &             - &             - &              - &              - &  - &              - &              - &              - &              - &             - \\
          SHOT* &             30.2	&34.1	&16.2&	22.6&	22.6&	32.4&	32.1&	45.5&	45.0&	34.5&	29.3&	47.8&	36.2&	\phz7.1&	\phz8.1&	29.6 \\
           % T3A &            - &              - &              - &              - &              - &             - &             - &             - &              - &              - &  - &              - &              - &              - &              - &             - \\
          TENT* &          29.5	&31.6	&17.6&	24.8	&27.2	&31.0&	32.4	&40.7	&35.0&	30.2&	26.6	&36.6&	29.3&	10.5&	12.4&	27.7  \\
        \midrule
 MATE-Standard* & 27.5&	29.4&	14.3	&\textbf{22.2}&	25.6&	40.8&	42.0&	\textbf{73.7}&	\textbf{63.2}&	45.1&	35.3&	\textbf{73.3}&	45.3&	\phz7.1	&\phz9.3&	36.9 \\

 MATE-Standard** & 	22.9 &	31.5	&13.4 &	17.0 &	24.3 &	34.1&	35.5 &	69.5 &	63.7 &	39.1 &	28.4 &	68.8 &	38.7 &	 7.7 &	 6.2 &	33.4 \\
\colrow
 SMART-PC-Standard & \textbf{27.5}&	\textbf{39.1}&	\textbf{19.3}	&21.5	&\textbf{29.8}	&\textbf{44.2}&	\textbf{48.9}&	68.3&	60.2	&\textbf{49.4}&	\textbf{45.4}&	70.1&	\textbf{49.1}&	\phz\textbf{8.4}&	\textbf{12.2}&	\textbf{39.6}\\

\midrule
 
   MATE-Online* &  33.0&	44.1&	13.3	&25.3	&29.1&	36.8&	37.7&	73.3&	65.2&	37.2&	31.3&	72.6&	40.6&	\phz7.2	&\phz7.4	&36.9 \\

   MATE-Online** &  29.4 &	33.9&	16.0&	25.5&	32.5&	34.8&	38.2&	69.4	&66.6&	40.6&	30.6&	70.2&	40.3&	8.1&	7.9	&36.3 \\

\colrow
SMART-PC-Online$\dagger$ & 39.4 &	\textbf{54.6} &	19.6 &	\textbf{40.1} &	40.3 &	54.4 &	55.9 &	\textbf{73.3} &	\textbf{69.5} &	55.1 &	50.1 &	\textbf{74.9} &	57.8 &	\phz6.9 &	\phz7.9 &	\textbf{46.7}\\
\colrow
 SMART-PC-Online & \textbf{42.3}	&53.7	&\textbf{23.8}&	37.5	&\textbf{41.6}&	\textbf{57.0}&	57.0&	71.1&68.8&	\textbf{57.3}&	\textbf{53.4}	&72.1&	\textbf{58.2}&	\phz\textbf{8.4}&	\phz\textbf{8.4}	& \textbf{47.4}\\
\midrule
\end{tabular}
}
\end{table*}

\begin{figure*}[!t]
\centering
\includegraphics[width=0.85\textwidth]{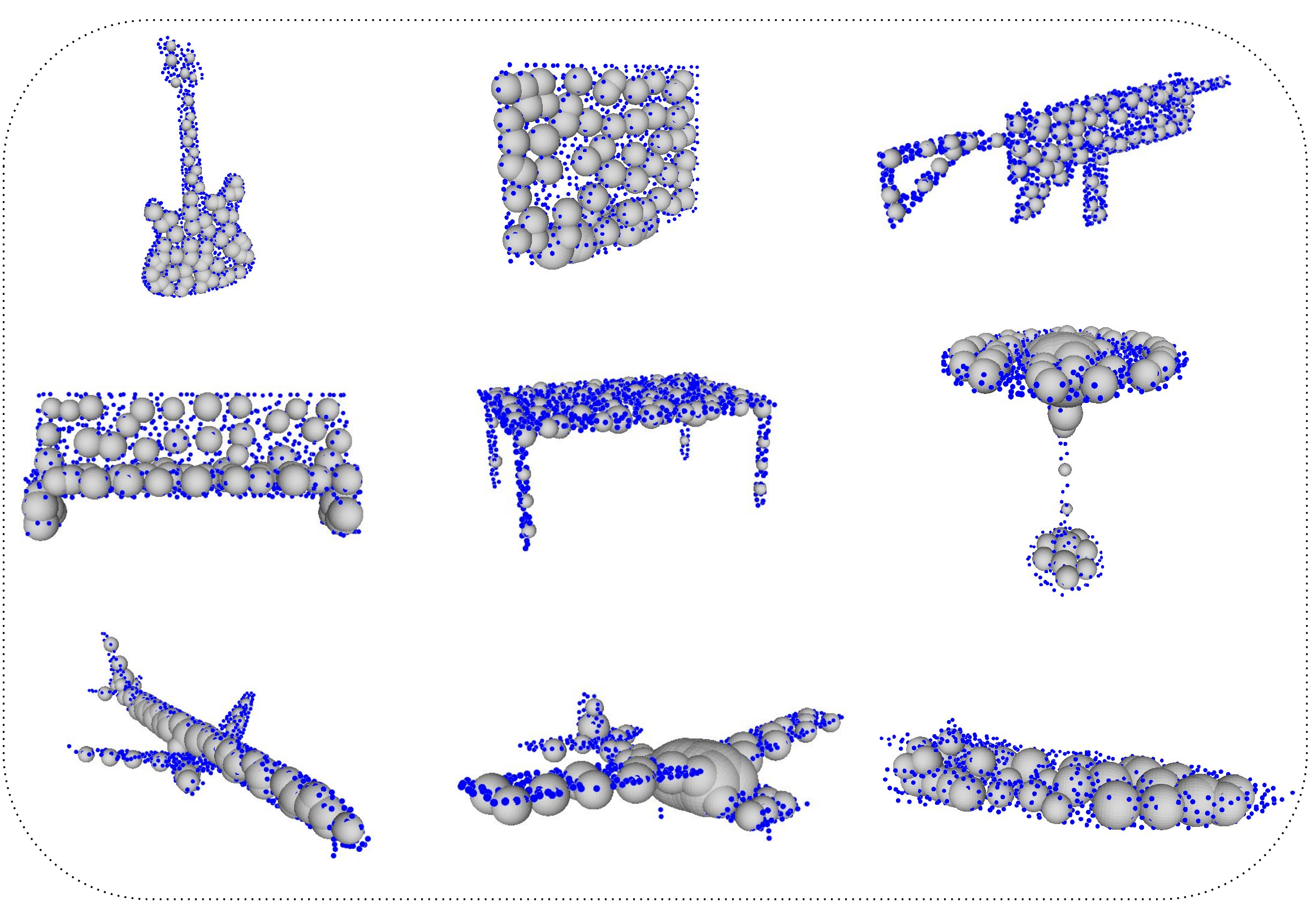}

\vspace{-12pt}
\caption{Visualization of 3D objects with original point clouds (blue dots) and their corresponding skeletal spheres.
}
\label{more_vis}
\end{figure*}

\section{More Visualizations}

\mypar{Skeletal Reconstruction.} In \cref{more_vis}, we present several 3D objects showcasing the original point clouds (blue dots) and their corresponding skeletal spheres. Each skeletal sphere is defined by a skeleton point (the center of the sphere) and its associated radius, which represents the local geometric structure around the skeleton point. The spheres collectively capture the essential geometric and structural features of the objects.

This visualization demonstrates the capability of our model to learn meaningful and compact representations of 3D shapes through skeletal abstraction. The skeleton effectively captures the underlying structure while filtering out high-frequency noise, enabling the model to focus on the fundamental geometry of the objects.
The diversity of shapes in this figure—ranging from guitars and airplanes to furniture—highlights the robustness of our approach in generalizing across different object classes.
These visualizations also provide insight into how the skeletal abstraction simplifies complex point cloud data into manageable representations, facilitating better performance in downstream tasks such as classification and adaptation under challenging conditions. This compact representation ensures that the geometric structure is preserved, even when the original point clouds are corrupted or noisy.
%%%%%%%%%%%%%%%%%%%%%%%%%%%%%%%%%%%%%%%%%%%%%%%%%%%%%%%%%%%%%%%%%%%%%%%%%%%%%%%
%%%%%%%%%%%%%%%%%%%%%%%%%%%%%%%%%%%%%%%%%%%%%%%%%%%%%%%%%%%%%%%%%%%%%%%%%%%%%%%

% \end{document}